\documentclass[11pt]{mystyle}

\usepackage{amsthm}
\usepackage{xcolor}
\usepackage{colortbl}
\usepackage{multirow}

\usepackage{mathtools,bm}
\usepackage{algorithm}
\usepackage{algpseudocode}
\usepackage{array,multirow}
\usepackage{longtable} %
\usepackage[numbers,sort&compress]{natbib}
\usepackage[nameinlink,capitalise,noabbrev]{cleveref}
\usepackage{tikz}
\usetikzlibrary{arrows.meta,positioning,fit,calc,shapes.geometric}

\usepackage[export]{adjustbox} %
\usepackage{fontawesome5} %
\usepackage{rotating} %
\usepackage{placeins} %
\usepackage{xspace} %
\usepackage{makecell} %
\usepackage{float} %
\usepackage{needspace} %

\makeatletter
\providecommand{\captionof}[1]{\def\@captype{#1}\caption}
\makeatother

\newcommand{\modellogo}[1]{%
  \IfFileExists{Figure/logo/#1.png}{\includegraphics[width=0.35cm,valign=c]{Figure/logo/#1.png}\,}{}%
}

\definecolor{S1Heat}{HTML}{4F8F87}
\definecolor{S2Heat}{HTML}{C08A4A}
\definecolor{TabGroup}{HTML}{EEF0DE}
\definecolor{TabRule}{HTML}{3D454E}
\definecolor{TabHead}{HTML}{898065}
\definecolor{TabBand}{HTML}{F7F7EF}
\definecolor{FamSem}{HTML}{4B79A5}
\definecolor{FamKnow}{HTML}{5F9668}
\definecolor{FamSafe}{HTML}{AD5563}
\definecolor{FamEvid}{HTML}{C0904A}
\definecolor{FamAgent}{HTML}{7A69A8}
\newcommand{\tabhead}[1]{\textcolor{white}{\bfseries #1}}
\definecolor{PipeOrange}{HTML}{EB6834}
\definecolor{PipeBlue}{HTML}{2A78D6}
\definecolor{PipeNavy}{HTML}{1A4F8F}
\definecolor{PipeSlate}{HTML}{64748B}
\definecolor{PipeGrey}{HTML}{E9ECF1}

\definecolor{HeaderBlue}{RGB}{220,235,247}
\definecolor{HeaderAmber}{RGB}{255,242,204}
\definecolor{HeaderRose}{RGB}{252,228,214}
\definecolor{HeaderViolet}{RGB}{233,221,243}
\definecolor{RuleGrey}{RGB}{170,176,188}
\definecolor{OursYellow}{RGB}{255,243,200}

\definecolor{TickGreen}{HTML}{4B9A72}
\definecolor{CrossRed}{HTML}{B0566F}
\definecolor{HalfAmber}{HTML}{C0904A}

\newcolumntype{C}{>{\columncolor{HeaderViolet!45}\centering\arraybackslash}m{1.95cm}}
\newcolumntype{B}{>{\columncolor{HeaderBlue!40}\centering\arraybackslash}m{1.63cm}}
\newcolumntype{J}{>{\columncolor{HeaderAmber!55}\centering\arraybackslash}m{1.78cm}}
\newcolumntype{M}{>{\columncolor{HeaderRose!45}\centering\arraybackslash}m{1.85cm}}

\theoremstyle{plain}

\theoremstyle{definition}

\theoremstyle{remark}

\crefname{assumption}{Assumption}{Assumptions}
\Crefname{assumption}{Assumption}{Assumptions}

\usepackage{authblk}

\makeatletter
\apptocmd{\ps@firststyle}{}{}{}
\makeatother

\tcbset{titlebox/.append style={left=4mm,right=4mm}}
\runningtitle{SO-Arena}

\title{SO-Arena\\[-2pt]
  {\fontsize{12}{13}\selectfont Benchmarking the Capability Spectrum and Reliability of System One Models}}
\DeclareRobustCommand{\paperauthorblock}{%
  {\normalfont\normalsize\bfseries Lichen Li$^{*}$, Ziping Dong$^{*}$, Sicheng Feng, Qi Li, Xinchao Wang$^{\dagger}$\par}%
}
\author{\paperauthorblock}
\date{}
\hypersetup{hidelinks}

\newcommand{\soneresourcerow}[3]{%
  \noindent\makebox[1.55em][c]{#1}\hspace{0.2em}%
  \makebox[8.4em][l]{\sffamily\bfseries #2:}%
  \if\relax\detokenize{#3}\relax\textcolor{TitleWine}{\itshape coming soon}%
  \else\href{#3}{\textcolor{TitleWine}{\nolinkurl{#3}}}\fi\par
}
\newcommand{\soneprojectlinks}{%
  \par\vspace{5pt}%
    {\normalfont\footnotesize\setlength{\parskip}{1pt}%
        \soneresourcerow{\faGlobe}{Homepage}{https://lichenlillc.github.io/SO-Arena}%
        \soneresourcerow{\includegraphics[height=1.18em,valign=c]{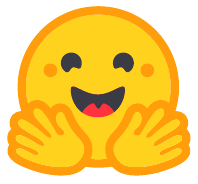}}{Hugging Face}{https://huggingface.co/SO-Arena}%
        \soneresourcerow{\faGithub}{GitHub}{https://github.com/holdrain/SO-Arena}%
        \noindent\makebox[1.55em][c]{\faEnvelope}\hspace{0.2em}%
        \makebox[8.4em][l]{\sffamily\bfseries Correspondence:}%
        \parbox[t]{\dimexpr\linewidth-10.15em\relax}{%
          \href{mailto:lichenli@u.nus.edu}{\textcolor{TitleWine}{\nolinkurl{lichenli@u.nus.edu}}},\quad
          \href{mailto:xinchao@nus.edu.sg}{\textcolor{TitleWine}{\nolinkurl{xinchao@nus.edu.sg}}}%
        }\par
    }%
}
\apptocmd{\abscontent}{\soneprojectlinks}{}{\PackageError{cle}{Could not insert project links}{Check the title-box definition in mystyle.cls.}}

\AtBeginDocument{\arrayrulecolor{TabRule}}
\begin{document}

\begin{abstract}
System One models promise fast, general-purpose decision-making without lengthy token generation. However, understanding their capability spectrum and decision reliability requires more than aggregate performance on current benchmarks. Existing evaluations often provide limited insight into the
breadth of model capabilities and the reliability of their decisions.
We introduce \textbf{SO-Arena}, a benchmark characterizing
the capability spectrum and decision reliability of System One
models across 11 domains and five capability families.
For the capability spectrum, SO-Arena offers
\textbf{evidence-grounded evaluation} through reliable sources
and manual verification of reference labels against model-visible
inputs; \textbf{fine-grained capability assessment} through
domain-specific difficulty criteria; and
\textbf{realistic interactive evaluation}, where models
iteratively execute actions in sandboxes. \textbf{Reasoning-intensive tasks} further enable
system one–two comparison.
Beyond accuracy, SO-Arena assesses \textbf{decision reliability} through two complementary measures: decision stability under
controlled option-set changes and missing-answer recognition when no valid option is available.
Evaluations across 21 models reveal that the benefits of scaling and additional reasoning vary substantially across domains. More importantly, strong overall performance does not
necessarily imply reliable decision-making: the best-performing System One model ranks only seventh in reliability. Together, these findings highlight the need to characterize capability
spectrum and decision reliability as complementary dimensions, providing a systematic view of the strengths and limitations of System One models.
\end{abstract}
\maketitle

\par\medskip
\noindent\begin{minipage}{\linewidth}
  \centering
  \includegraphics[width=\linewidth]{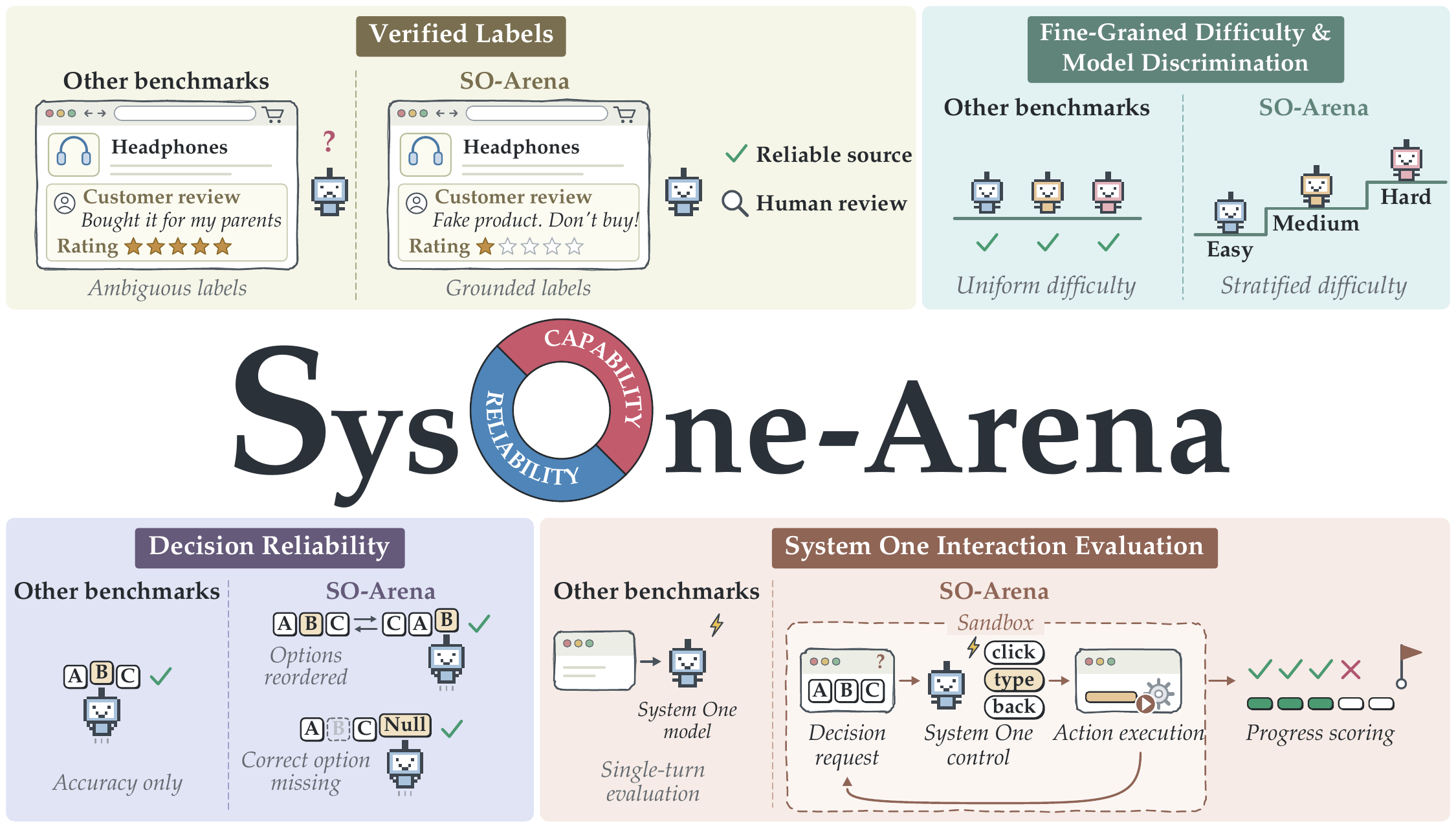}
  \captionsetup{skip=6pt,font=footnotesize}
  \captionof{figure}{\textbf{Highlights of SO-Arena.} SO-Arena evaluates System One models on capability and reliability. Compared with other benchmarks, it verifies reference labels, stratifies tasks by difficulty to discriminate between models, evaluates long-horizon agentic ability in sandboxes built for System One, and tests reliability under reordered options and a missing correct option.}
  \label{fig:teaser}
\end{minipage}
\enlargethispage{32pt}
\vfill
\begingroup
\normalfont\footnotesize\color{black}
\footnoterule
\noindent\hspace*{1.8em}$^{*}$Equal contribution.\quad $^{\dagger}$Corresponding author.\par
\endgroup
\clearpage

\newlength{\frontpanelheight}
\setlength{\frontpanelheight}{0.5482\linewidth}
\noindent\begin{minipage}{\linewidth}
  \centering
  \begin{minipage}[t]{0.45\linewidth}
    \centering\vspace{0pt}
    \includegraphics[height=\frontpanelheight]{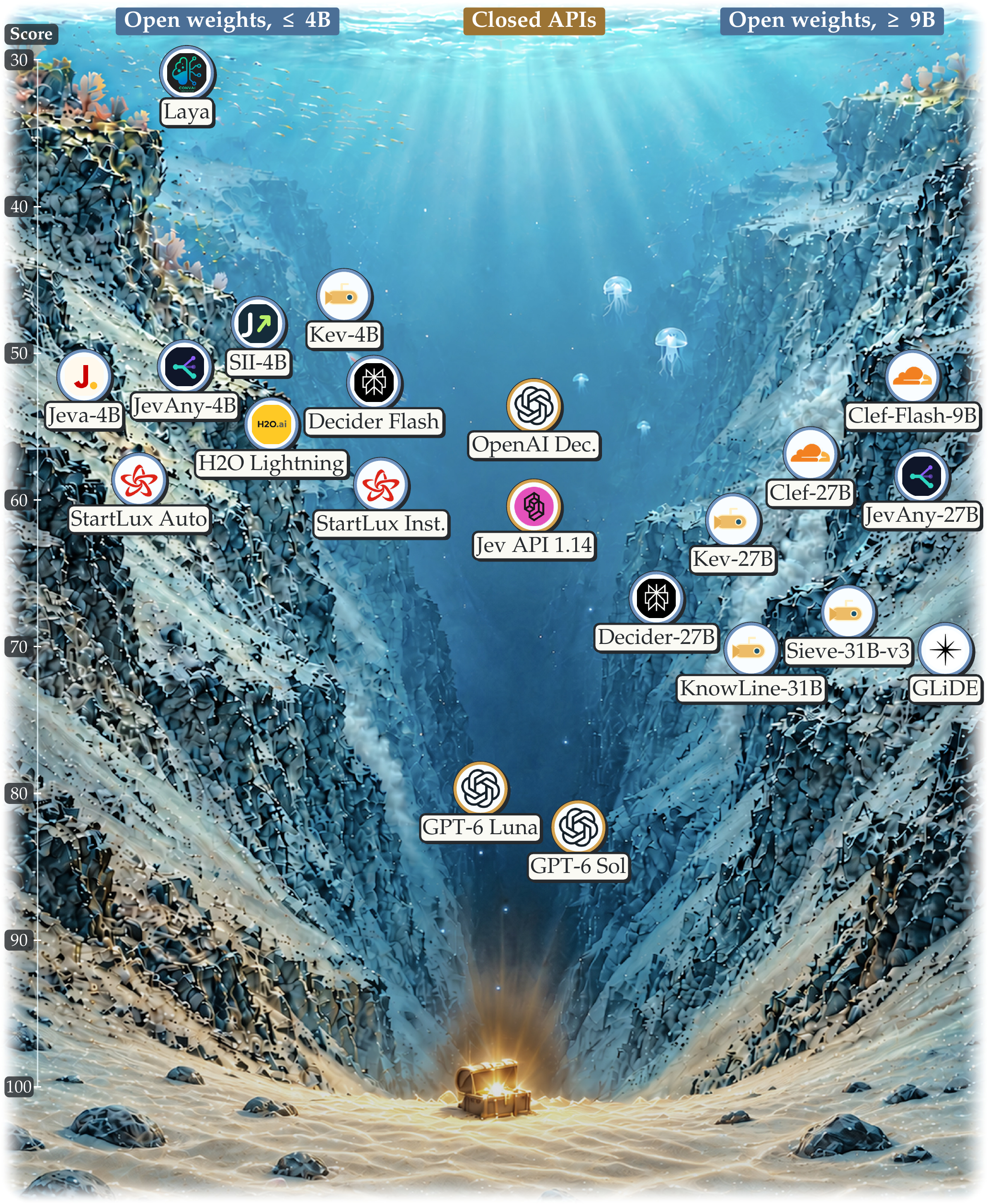}
    \par\smallskip{\footnotesize\textbf{(a) Deep-dive ranking}}
  \end{minipage}\hfill
  \begin{minipage}[t]{0.545\linewidth}
    \centering\vspace{0pt}
    \includegraphics[height=\frontpanelheight]{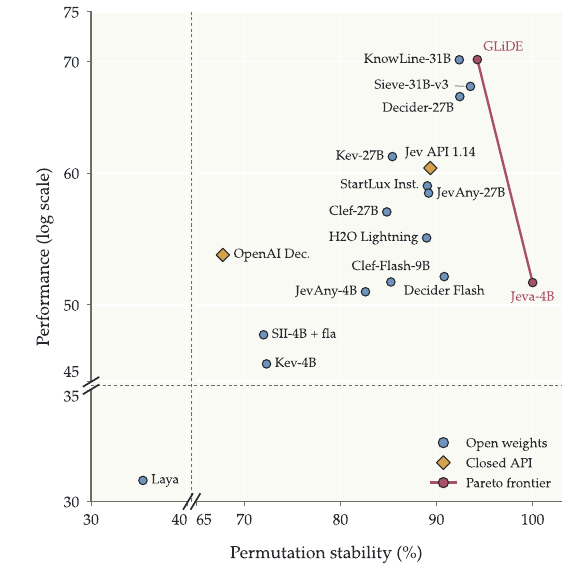}
    \par\smallskip{\footnotesize\textbf{(b) Accuracy and stability}}
  \end{minipage}
  \captionsetup{skip=6pt,font=footnotesize}
  \captionof{figure}{\textbf{SO-Arena at a glance.} (a) Leaderboard scores for 21 models; greater depth indicates a higher score, and models are grouped horizontally by size and access. (b) Performance (the leaderboard score) versus stability under option reordering ($100S$) for the 19 models with stability measurements; the red line marks the Pareto frontier.}
  \label{fig:front-results}
\end{minipage}

\par\bigskip
\section{Introduction}
\label{sec:introduction}

System Two reasoning has substantially advanced language model capabilities in mathematics, coding, and science~\citep{deepseek2025v32,openai2026gpt6astra,anthropic2026opus55,google2025gemini3pro,kong2026geo,liu2026timesjev}. 
These gains often come at a higher inference cost, as models generate additional reasoning tokens before answering~\citep{feng2025efficient,openaio1,guo2025deepseekr1,fang2025thinkless,chen2025verithinker}.
Yet many routine decisions in software workflows, such as routing requests, selecting tool actions, and verifying task completion, can be made without generating explicit reasoning traces.
System One models~\citep{almeida2026jev,wei2026jevany,palmer2026kev27b,wei2026jevany,fastino2026glide,pelaai2026knowline,zhuang2026startlux} such as Jev~\citep{almeida2026jev} predict a distribution over predefined outcomes in a single forward pass, substantially reducing inference latency and providing structured outputs that downstream software can use directly.

Recent benchmarks evaluate System One models across a range of decision tasks and settings~\citep{jevbench,brida2026reflexbench,
tateno2026s1mb,alhashimi2026systemonebench,chen2026system1arena}. Building on these evaluations, we argue that a benchmark designed to characterize the capabilities and reliability of System One models should provide trustworthy reference labels supported by model-visible evidence, distinguish performance across increasing
task complexity, evaluate models both as actors and verifiers in interactive settings, and assess decision reliability under controlled changes to the available options.

To address these requirements, we introduce \textbf{SO-Arena}, a benchmark comprising 2{,}919 tasks across eleven domains organized into five capability families: semantic understanding, knowledge and reasoning, safety and policy compliance, evidence assessment, and agent task completion. The benchmark brings together static and environment-grounded tasks under a unified decision interface, with performance evaluated at the level of the original task. We curate tasks from traceable sources, manually verify that reference labels are supported by the information available to the model, and keep evaluation references separate from model inputs.
To characterize the capability spectrum beyond aggregate accuracy, we organize nine domains into difficulty tiers based on domain-specific criteria for task complexity. We report performance across these tiers and construct a capability leaderboard that weights all eleven domains equally while emphasizing the most demanding tasks. In agentic settings, we further distinguish between action selection and completion verification. The former evaluates whether models can accomplish goals through sequential decisions in web and tool environments, while the latter assesses whether they can correctly judge recorded executions, including cases where identical final goal states conceal different compliance with intermediate requirements. We also incorporate reasoning-intensive tasks to compare direct System One decisions with explicit System Two reasoning under matched task conditions. Beyond task performance, we examine decision reliability under changes to the available options. Specifically, we test whether models maintain consistent decisions when options are reordered and recognize the absence of a valid answer when the correct option is removed. These two aspects are combined into a separate reliability score, allowing us to assess whether strong capability performance is accompanied by reliable decision-making.

We evaluate 21 models on SO-Arena in terms of capability, inference latency, and decision reliability. The two System Two references lead the capability leaderboard, while
GLiDE~\cite{fastino2026glide} and KnowLine-31B~\cite{pelaai2026knowline} achieve the highest scores among System One models, illustrating the remaining performance gap despite the latter's substantial speed advantage. Reliability, however, does not follow the same ranking: JevAny-27B~\cite{wei2026jevany} achieves the highest reliability
among the evaluated System One models despite ranking eleventh in capability. Beyond these aggregate results, we find that performance declines with increasing task difficulty, particularly in agentic settings, and that the benefits of stronger backbones
and additional reasoning vary substantially across domains.

These findings demonstrate the importance of characterizing the capability spectrum and decision reliability of System One models as complementary dimensions. 

Our main contributions are summarized as follows:

\begin{itemize}
    \item
    We introduce SO-Arena, comprising 2,919 tasks across 11
    domains and five capability families. We establish a
    traceable evaluation process and manually verify that
    reference labels are supported by the information
    available to the model.

    \item
    We introduce domain-specific difficulty tiers to
    distinguish model capabilities and extend evaluation
    beyond static judgments to interactive execution and
    history-dependent task verification.

    \item 
    We develop a separate reliability assessment combining decision consistency under option reordering and the ability to recognize missing correct answers, revealing aspects of decision-making overlooked by conventional accuracy metrics.

    \item
    We evaluate 21 models to reveal substantial capability
    differences across domains and difficulty levels, uneven
    benefits from model capacity and additional reasoning, and a notable divergence
    between capability and reliability rankings.
\end{itemize}

\section{Preliminaries}
\label{sec:background}
\label{sec:background-decision-models}

We use \emph{System One decision models} to denote models that
map a supplied state and a decision specification to structured
predictions over admissible outcomes. Following existing decision-task formulations~\citep{duan2026general,almeida2026jev},
we represent an instance as
\begin{equation}
    x = (s,q,\mathcal{C}),
    \label{eq:decision-instance}
\end{equation}
where $s$ contains the supplied information, $q$ specifies the
decision objective, and
$\mathcal{C}=\{c_1,\ldots,c_K\}$ defines the admissible outcomes
and their descriptions. The supplied state may contain text,
structured records, retrieved passages, or interaction histories.
A decision model returns a distribution
$p_{\theta}(\cdot\mid x)$ satisfying
\begin{equation}
    p_{\theta}(c_k\mid x)\geq 0,
    \qquad
    \sum_{k=1}^{K}p_{\theta}(c_k\mid x)=1.
    \label{eq:decision-distribution}
\end{equation}
This distribution can be obtained through a learned decision
head or a readout from language-model logits, among other
implementations~\citep{palmer2026kev4b,geuter2026innerjev}.
Across these implementations, the decision type specifies the structure
of the outcome space over which the model assigns probabilities.
We consider three output formats supported by the Jev-style
interface~\citep{typesafePrimitives}:

\begin{itemize}
    \item \textbf{Categorical selection (\texttt{choice}).}
    The outcomes are explicitly specified alternatives without
    an imposed ordinal structure. The predicted outcome is
    \begin{equation}
        \hat{y}(x)
        =
        \operatorname*{arg\,max}_{c\in\mathcal{C}}
        p_{\theta}(c\mid x).
        \label{eq:categorical-prediction}
    \end{equation}

    \item \textbf{Binary judgment (\texttt{noul}, the interface's yes/no question type).}
    The outcome space is $\{0,1\}$, and the model returns
    $p_{\theta}(1\mid x)$, interpreted as the probability
    assigned to the affirmative outcome.

    \item \textbf{Ordinal assessment (\texttt{score}).}
    The outcomes are ordered descriptive levels with grades
    $g(c_k)$ on the source's native scale. The predicted grade
    is that of the most probable level,
    $\hat{g}(x)=g(\hat{y}(x))$.
    An ordinal ordering alone does not establish equal semantic
    distances between levels, so we compare grades only through
    a bounded credit for adjacent levels.
\end{itemize}

These formats describe the output space. They do not determine
the task domain or the reasoning required to answer a question.

\vspace{-1mm}

\begin{figure}[t]
  \centering
  \includegraphics[width=\linewidth]{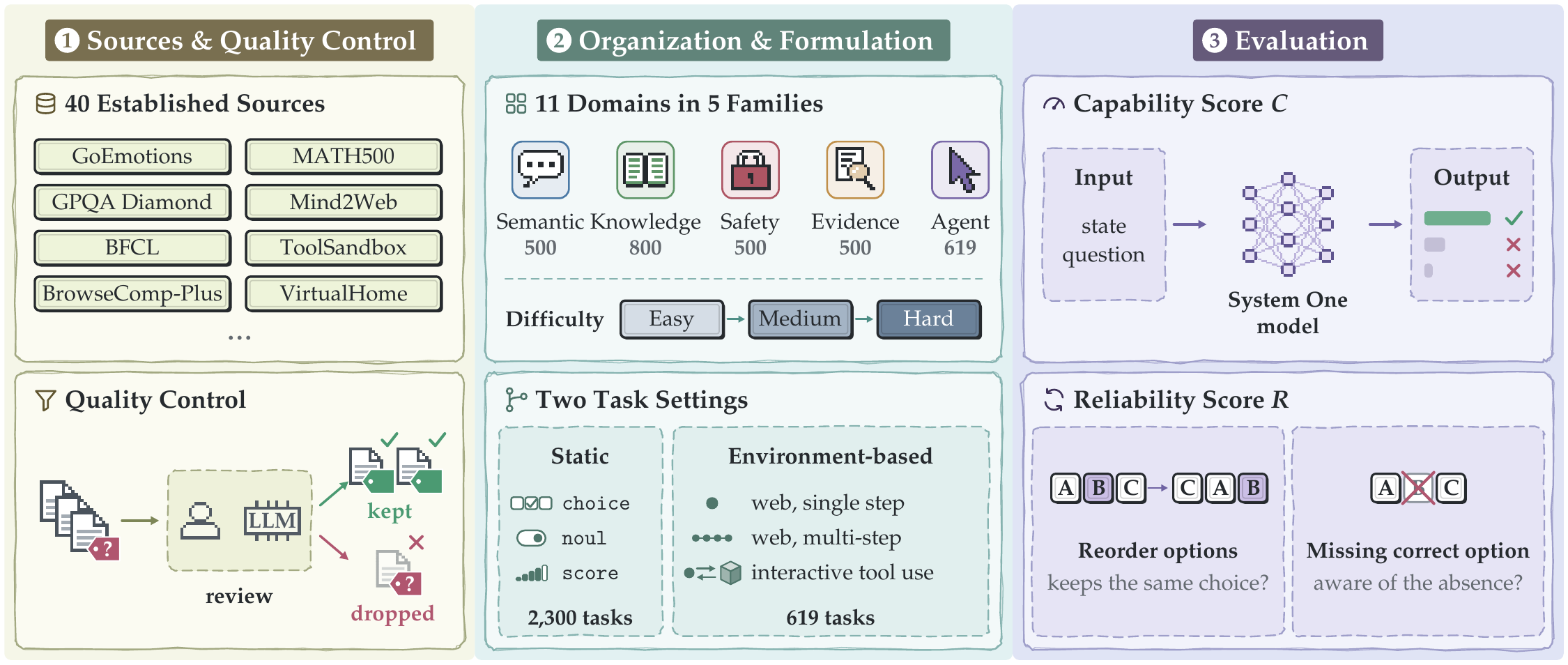}
  \captionsetup{skip=6pt,font=footnotesize}
  \caption{\textbf{Overview of SO-Arena.} (1) Tasks are drawn or derived from 40 established public sources; items with questionable labels are dropped after human and LLM review. (2) The 2{,}919 tasks span 11 domains in five capability families, nine of which have domain-relative difficulty tiers, and are posed as static or environment-based tasks. (3) The capability score $C$ averages the hardest tier of each domain with equal domain weights, and the reliability score $R$ combines the option-reordering and missing-correct-option tests.}
  \label{fig:bench-overview}
\end{figure}

\section{Benchmark Design}
\label{sec:bench}
This section describes how we construct and evaluate our benchmark for System One decision-making. We first collect and derive tasks from public sources and control their quality (\cref{subsec:sources-curation}). We then organize these tasks by domain and, where applicable, difficulty tier, and formulate static and environment-based tasks using a common decision interface (\cref{subsec:task organization}). Given these task formulations, we define metrics for capability across domains and decision reliability (\cref{subsec:evaluation-dimensions}). ~\cref{fig:bench-overview} summarizes the pipeline.

\subsection{Data Sources and Quality Control}
\label{subsec:sources-curation}
We draw on 40 distinct public datasets, benchmarks, and environments
across 11 domains. Representative sources include GoEmotions~\citep{demszky2020goemotions} for
emotion understanding, MATH500~\citep{lightman2023verify} for mathematics,
GPQA Diamond~\citep{rein2023gpqa} for scientific reasoning,
Mind2Web~\citep{deng2023mind2web} for web interaction, and
BFCL~\citep{patil2025bfcl} and ToolSandbox~\citep{lu2024toolsandbox}
for tool use. We convert the source data to our decision formats or derive new tasks from them, such as completion checks on BrowseComp-Plus~\citep{chen2025browsecompplus} search runs and VirtualHome~\citep{puig2018virtualhome} trajectories. For quality control, we keep source labels as references, review them manually and with LLMs, and exclude items with questionable references. We record each task's source, version, and original identifier where applicable, and flag model-derived references in the metadata. 

\subsection{Task Organization, Formulation, and Evaluation Settings}
\label{subsec:task organization}

\begin{figure}[t]
  \centering
  \begin{minipage}[t]{0.53\linewidth}
    \centering\vspace{0pt}
    \includegraphics[width=\linewidth]{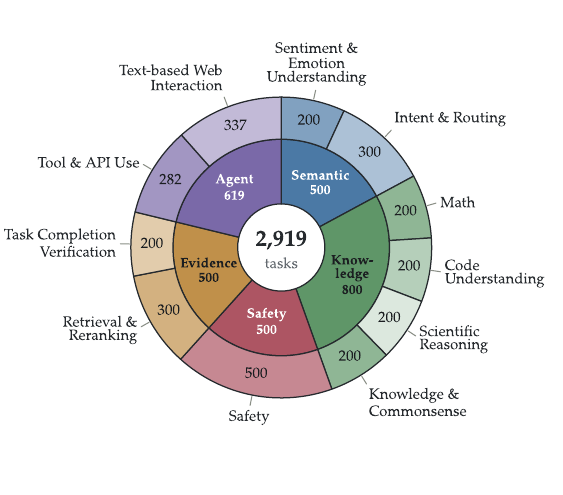}
    \par\smallskip{\footnotesize\textbf{(a) Families and domains}}
  \end{minipage}\hfill
  \begin{minipage}[t]{0.46\linewidth}
    \centering\vspace{0pt}
    \includegraphics[width=\linewidth]{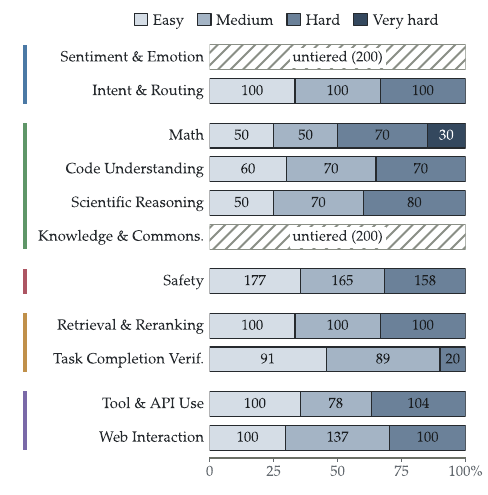}
    \par\smallskip{\footnotesize\textbf{(b) Difficulty tiers within each domain}}
  \end{minipage}
  \captionsetup{skip=6pt,font=footnotesize}
  \caption{\textbf{Composition of the benchmark.} (a) Natural tasks by capability family (inner ring) and domain (outer ring). (b) Share of each domain's tasks in each difficulty tier, with task counts on the bars; side rules mark the families in (a). Tiers are defined within each domain and are not calibrated across domains.}
  \label{fig:bench-composition}
\end{figure}

We group the 11 domains into five capability families covering
semantic understanding, knowledge and reasoning, safety and policy
compliance, evidence assessment, and agent task completion.
Of the 11 domains, nine have difficulty tiers. The remaining two,
Sentiment \& Emotion Understanding and Knowledge \& Commonsense,
are untiered because their tasks offer limited scope for systematically varying decision complexity. In the tiered domains, difficulty is relative to each domain and reflects the demands of its tasks. For example, Math uses source difficulty levels or the number of calculation steps, depending on the source. Task Completion Verification distinguishes direct state checks, judgments that require integrating
evidence, and judgments that depend on the execution history.
\Cref{fig:bench-composition}(b) shows the distribution of tasks across
tiers, and \Cref{app:construction} specifies the criteria for each tiered domain.

To evaluate these capabilities in settings that reflect real-world
applications, we consider how the context of a task shapes the model's decision. Some decisions can be made solely from a fixed input, while others are made from the current state of an environment, such as a website or a sandbox with tools; in sequential tasks, each decision changes the state for the next, so later decisions depend on earlier ones being correct. We therefore organize the benchmark into two task settings, static and environment-based.

\textbf{Static Tasks.} In static tasks, the model judges a fixed input. We convert each source task into one of the three formats and keep the source's answer as the reference. For \texttt{choice}, multiple-choice questions are used as they are, and classification tasks offer the source's labels as options. Open-answer math problems have no options, so we build distractors from wrong answers of a base model, intermediate results of the solution, and rule-based near misses. For \texttt{noul}, binary judgments are posed as yes/no questions, and a multi-label task becomes one yes/no question per candidate label. For \texttt{score}, ordinal ratings keep the source's native scale. For tasks we build ourselves, we establish the reference by execution, verified construction, or annotation (\Cref{app:construction}). We exclude tasks that require free-text responses.

\textbf{Environment-Based Tasks.} In environment-based tasks, the model assesses or selects actions given a task goal and an observation from a web or tool environment. Single-step web tasks show one recorded page state, on which the model judges whether a proposed action advances the goal or selects the best candidate action. Sequential tasks require a series of decisions, each a choice from a menu of candidate actions. In multi-step web tasks, the model acts at successive recorded states along a reference route that replay has verified to complete the task.\footnote{Alternative action orders and harmless detours are also accepted when replay verifies that they still permit completion.} In interactive tool-use episodes, the
BFCL~\citep{patil2025bfcl} or ToolSandbox~\citep{lu2024toolsandbox}
simulator executes each selected action and returns the observation on which the next decision is made. We score each episode based on the corresponding benchmark's success criteria (\Cref{app:tool-scoring}).

The resulting benchmark comprises 2{,}300 static tasks and 619
environment-based tasks. The latter include 337 web tasks
(100 single-step judgments and 237 reference-route tasks) and
282 interactive tool-use episodes. \Cref{fig:bench-composition}(a)
shows the distribution of tasks across capability families and domains.
Table~\ref{tab:bench-composition} summarizes the sources, task counts, difficulty tiers, and decision formats for each domain.

\FloatBarrier
\subsection{Evaluation Metrics}
\label{subsec:evaluation-dimensions}

\subsubsection{Capability across domains.}
We use the original task as the unit of scoring and weight domains
equally so that larger domains do not dominate the leaderboard.
Let $c_i(m)\in[0,1]$ denote model $m$'s credit on task $i$.
For domain $d$, let $\mathcal{T}_d$ contain the hardest tier, both the
hard and very hard tiers for Math, or all tasks for untiered domains.
Focusing on the upper tiers makes the leaderboard more sensitive to
performance differences on demanding tasks. The capability score is
\begin{equation}
    C(m)=\frac{100}{D}\sum_{d=1}^{D}
    \frac{1}{|\mathcal{T}_d|}\sum_{i\in\mathcal{T}_d}c_i(m),
    \label{eq:capability-score}
\end{equation}
where $D$ is the number of domains.

\subsubsection{Reliability}
\paragraph{Option order stability.}
A reliable model should pick the same option however the options are
ordered. We show each item $i$ in its original order and in 12
reorderings, and identify the model's answer by the option's content,
not its position. $A(m)$ is the share of reorderings in which model $m$
keeps its original answer; $S(m)$ rescales it so that random picks score
0 and full agreement scores 1:
\begin{equation}
    S(m)=\frac{A(m)-p_e}{1-p_e},
    \qquad
    A(m)=\frac{1}{12M}\sum_{i=1}^{M}\sum_{r=1}^{12}\mathbf{1}\{a_i^{(r)}=a_i^{(0)}\},
    \qquad
    p_e=\frac{1}{M}\sum_{i=1}^{M}\frac{1}{K_i},
    \label{eq:order-stability}
\end{equation}
where $a_i^{(0)}$ and $a_i^{(r)}$ are the answers in the original order
and in reordering $r$, $M$ is the number of items, and $p_e$ is the
agreement of random picks among item $i$'s $K_i$ options. $S$ measures
consistency, not correctness.

\paragraph{Recognition of missing correct options.}
A model should assign a higher probability to the absence of a correct
answer (\textit{Null}) when none of the offered options is correct. We define this as
$p_{\mathrm{null}}=1-p(\mathrm{inc})$, where $p(\mathrm{inc})$ is the
model's estimated probability that at least one correct option is included (\Cref{app:configs}).
We show each item twice, as it is and with its correct option removed,
and use $p_{\mathrm{null}}$ to predict which copies lack a correct option\footnote{Exception: StartLux Instant is scored with their own null output}.
Null discrimination is
\begin{equation}
    N(m)=2\,\operatorname{AUC}(m)-1,
    \label{eq:null-discrimination}
\end{equation}
where $\operatorname{AUC}(m)$ is the area under the ROC curve of this
prediction over all copies: the probability that a random copy without a
correct option receives a higher $p_{\mathrm{null}}$ than a random copy
with one, with ties counted as half. $N$ is 0 at chance and 1 when every
copy without a correct option scores higher.

\paragraph{Combined reliability.}
Reliable decisions should be stable under reordering and sensitive to
whether a correct option exists. We combine the two components using
their geometric mean, with a positive score only when both exceed chance:
\begin{equation}
    R(m)=
    \begin{cases}
        \sqrt{S(m)N(m)}, & S(m)>0\ \text{and}\ N(m)>0,\\
        0, & \text{otherwise}.
    \end{cases}
    \label{eq:reliability-score}
\end{equation}
We report the reliability score $R(m)$ as a percentage alongside
the capability score $C(m)$ in \Cref{fig:front-results}.

\begin{table}[t]
  \centering
  \captionsetup{font=footnotesize}
  \caption{\textbf{The 21 evaluated models.} Open weights: \textcolor{TickGreen}{\ding{51}}~available, \textcolor{CrossRed}{\ding{55}}~closed API. Reasoning: whether the model reasons before answering; \emph{auto} lets the model decide for each request. Releases and the settings of our latency and reliability runs are listed in \Cref{tab:config-settings}.}
  \label{tab:configs}
  \fontsize{8.2}{9.8}\selectfont
  \setlength{\tabcolsep}{2.5pt}
  \begin{tabular}{@{}llrcc@{\hskip 14pt}llrcc@{}}
    \rowcolor{TabHead}
    \tabhead{Model} & \tabhead{Developer} & \tabhead{Size} & \tabhead{\makecell{Open\\weights}} & \tabhead{Reasoning} & \tabhead{Model} & \tabhead{Developer} & \tabhead{Size} & \tabhead{\makecell{Open\\weights}} & \tabhead{Reasoning} \\
    GLiDE~\citep{fastino2026glide} & Fastino & 27B & \textcolor{TickGreen}{\ding{51}} & No & Decider Flash~\citep{perplexity2026deciderflash} & Perplexity & 4B & \textcolor{TickGreen}{\ding{51}} & No \\
    \rowcolor{TabBand}KnowLine-31B~\citep{pelaai2026knowline} & PelaAI & 31B & \textcolor{TickGreen}{\ding{51}} & No & JevAny-4B~\citep{wei2026jevany} & SimpleJev & 4B & \textcolor{TickGreen}{\ding{51}} & No \\
    Sieve-31B-v3~\citep{sthanika2026sieve} & Sthanika AI & 31B & \textcolor{TickGreen}{\ding{51}} & No & SII-4B~\citep{duan2026general} & SII & 4B & \textcolor{TickGreen}{\ding{51}} & No \\
    \rowcolor{TabBand}Decider-27B~\citep{perplexity2026decider} & Perplexity & 27B & \textcolor{TickGreen}{\ding{51}} & No & Kev-4B~\citep{palmer2026kev4b} & J.~Palmer & 4B & \textcolor{TickGreen}{\ding{51}} & No \\
    Kev-27B~\citep{palmer2026kev27b} & J.~Palmer & 27B & \textcolor{TickGreen}{\ding{51}} & No & Jeva-4B~\citep{li2026jeva} & NUS & 4B & \textcolor{TickGreen}{\ding{51}} & No \\
    \rowcolor{TabBand}JevAny-27B~\citep{wei2026jevany} & SimpleJev & 27B & \textcolor{TickGreen}{\ding{51}} & No & Laya~\citep{convai2026laya} & ConvAI Innovations & 0.4B & \textcolor{TickGreen}{\ding{51}} & No \\
    StartLux Auto~\citep{zhuang2026startlux} & StartLux & 4B & \textcolor{TickGreen}{\ding{51}} & Auto & GPT-6 Sol~\citep{openai2026solandluna} & OpenAI & -- & \textcolor{CrossRed}{\ding{55}} & Medium effort \\
    \rowcolor{TabBand}StartLux Instant~\citep{zhuang2026startlux} & StartLux & 4B & \textcolor{TickGreen}{\ding{51}} & No & GPT-6 Luna~\citep{openai2026solandluna} & OpenAI & -- & \textcolor{CrossRed}{\ding{55}} & Medium effort \\
    Clef-27B~\citep{cloudflare2026clef} & Cloudflare & 27B & \textcolor{TickGreen}{\ding{51}} & No & Jev API 1.14~\citep{almeida2026jev} & TypeSafe & -- & \textcolor{CrossRed}{\ding{55}} & Not disclosed \\
    \rowcolor{TabBand}H2O Lightning~\citep{h2oai2026lightning} & H2O.ai & 4B & \textcolor{TickGreen}{\ding{51}} & No & OpenAI Decisions~\citep{openai2026decisions} & OpenAI & -- & \textcolor{CrossRed}{\ding{55}} & Not disclosed \\
    Clef-Flash-9B~\citep{cloudflare2026clefflash} & Cloudflare & 9B & \textcolor{TickGreen}{\ding{51}} & No & & & & & \\
    \bottomrule
  \end{tabular}
\end{table}

\suppressfloats[t]
\section{Experiments}
\label{sec:exp}
System One models decide much faster than System Two models, but it remains unclear whether, and in which domains, they lose performance. Meanwhile, their speed suits high-volume, repeated daily tasks such as classification, routing, and real-time responses, where decisions must also be consistent. Our experiments therefore measure how well and how fast current System One models decide and how reliable their decisions are, across 11 major domains. \Cref{sec:exp-main} reports the main results; \Cref{sec:discussion} examines where System One models match a System Two reference, what narrows the gap where they do not, and which factors shape reliability.

\subsection{Main Results}
\label{sec:exp-main}
We evaluate 21 models shown in \Cref{tab:configs}: 16 open-weight System One models from 0.4B to 31B, one model that switches automatically between System One and Two (StartLux Auto), two closed decision APIs (Jev API 1.14 and OpenAI Decisions), and two closed System Two references (GPT-6 Luna and Sol). Each model runs in the settings documented by its developer\footnote{With two exceptions: Laya reads every option in full length, without its default cut at 48 tokens; our latency experiments run SII-4B with flash-linear-attention, shown as SII-4B + fla in \Cref{fig:latency}.} (\Cref{app:configs}). We evaluate them along two dimensions. Performance is paired with latency, since speed is what System One models are built for (\Cref{sec:exp-perf}). Reliability asks whether decisions hold when the options are reordered or the correct option is absent (\Cref{sec:exp-reliability}).

\subsubsection{Performance and Latency}
\label{sec:exp-perf}
\begin{figure}[t]
  \centering
  \includegraphics[width=\linewidth]{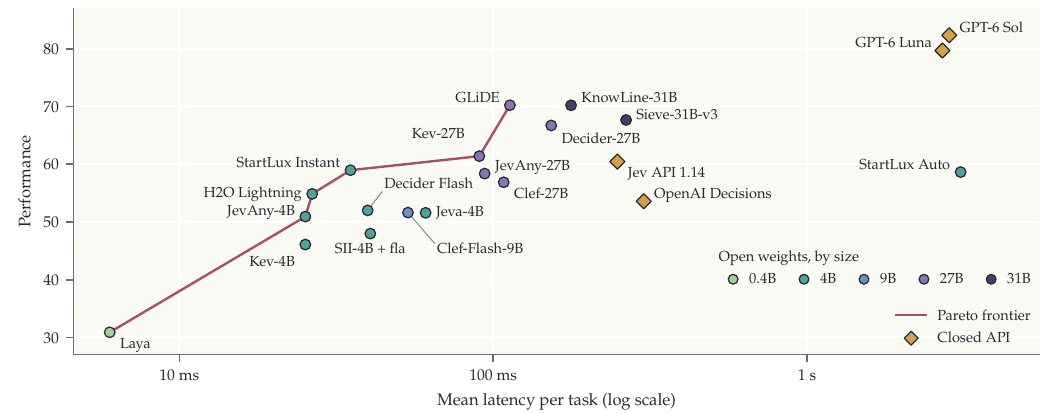}
  \captionsetup{font=footnotesize}
  \caption{\textbf{Performance against latency.} Performance is given by the leaderboard score (\Cref{eq:capability-score}). The Pareto frontier is formed by the System One models with optimal performance at each latency. Open-weight models run on one H100 GPU each, one task at a time. Closed APIs are timed end to end over the network (\Cref{app:latency}).}
  \label{fig:latency}
\end{figure}

As shown in \Cref{tab:tier-scores} and \Cref{fig:latency}, System Two models still lead in overall performance: GPT-6 Sol and Luna outperform every System One model, and the strongest System One models, GLiDE and KnowLine-31B, trail Luna by about ten points. \Cref{sec:analysis-domains} shows that this lead does not hold in every domain. In exchange, System One models are much faster. Every open-weight System One model answers within 0.3 seconds per task on one H100 GPU, tens of times faster than GPT-6 Sol and Luna (\Cref{fig:latency}).

\begin{table}[t]
  \centering
  \captionsetup{font=footnotesize}
  \caption{\textbf{Scores by domain and difficulty tier.} E, M, H, and V denote the easy, medium, hard, and very hard tiers; All marks untiered domains. Score is the leaderboard score (\Cref{eq:capability-score}). Rows are grouped by access and size and sorted by Score within each group; the two System Two references are shaded orange. Darker is higher; bold is the best in each column. Tiers are not calibrated across domains.}
  \label{tab:tier-scores}
  \fontsize{8.2}{9.8}\selectfont
  \setlength{\tabcolsep}{1.2pt}
  \renewcommand{\arraystretch}{1.12}
  \begin{tabular}{@{}l r !{\hskip 2.2pt}c !{\hskip 2.2pt}ccc !{\hskip 2.2pt}cccc !{\hskip 2.2pt}ccc !{\hskip 2.2pt}ccc !{\hskip 2.2pt}c !{\hskip 2.2pt}ccc !{\hskip 2.2pt}ccc !{\hskip 2.2pt}ccc !{\hskip 2.2pt}ccc !{\hskip 2.2pt}ccc@{}}
\multicolumn{1}{c}{\cellcolor{TabHead}} & \multicolumn{1}{c}{\cellcolor{TabHead}} & \multicolumn{1}{c}{\cellcolor{FamSem}\tabhead{Sent.}} & \multicolumn{3}{c}{\cellcolor{FamSem}\tabhead{Intent}} & \multicolumn{4}{c}{\cellcolor{FamKnow}\tabhead{Math}} & \multicolumn{3}{c}{\cellcolor{FamKnow}\tabhead{Code}} & \multicolumn{3}{c}{\cellcolor{FamKnow}\tabhead{Science}} & \multicolumn{1}{c}{\cellcolor{FamKnow}\tabhead{Know.}} & \multicolumn{3}{c}{\cellcolor{FamSafe}\tabhead{Safety}} & \multicolumn{3}{c}{\cellcolor{FamEvid}\tabhead{Retrieval}} & \multicolumn{3}{c}{\cellcolor{FamEvid}\tabhead{Verification}} & \multicolumn{3}{c}{\cellcolor{FamAgent}\tabhead{Tools}} & \multicolumn{3}{c}{\cellcolor{FamAgent}\tabhead{Web}} \\[-0.5pt]
\multicolumn{1}{c}{\cellcolor{TabHead}\multirow{-2}{*}{\tabhead{Model}}} & \multicolumn{1}{c}{\cellcolor{TabHead}\multirow{-2}{*}{\tabhead{Score}}} & \cellcolor{FamSem!22}All & \cellcolor{FamSem!22}E & \cellcolor{FamSem!22}M & \cellcolor{FamSem!22}H & \cellcolor{FamKnow!22}E & \cellcolor{FamKnow!22}M & \cellcolor{FamKnow!22}H & \cellcolor{FamKnow!22}V & \cellcolor{FamKnow!22}E & \cellcolor{FamKnow!22}M & \cellcolor{FamKnow!22}H & \cellcolor{FamKnow!22}E & \cellcolor{FamKnow!22}M & \cellcolor{FamKnow!22}H & \cellcolor{FamKnow!22}All & \cellcolor{FamSafe!22}E & \cellcolor{FamSafe!22}M & \cellcolor{FamSafe!22}H & \cellcolor{FamEvid!22}E & \cellcolor{FamEvid!22}M & \cellcolor{FamEvid!22}H & \cellcolor{FamEvid!22}E & \cellcolor{FamEvid!22}M & \cellcolor{FamEvid!22}H & \cellcolor{FamAgent!22}E & \cellcolor{FamAgent!22}M & \cellcolor{FamAgent!22}H & \cellcolor{FamAgent!22}E & \cellcolor{FamAgent!22}M & \cellcolor{FamAgent!22}H \\
\midrule
\multicolumn{32}{@{}l}{\textit{Closed-source APIs}} \\
\cellcolor{S2Heat!15}GPT-6 Sol & \cellcolor{S2Heat!15}82.3 & \cellcolor{S2Heat!39}88 & \cellcolor{S2Heat!52}\textbf{100} & \cellcolor{S2Heat!31}78 & \cellcolor{S2Heat!34}\textbf{82} & \cellcolor{S2Heat!52}\textbf{100} & \cellcolor{S2Heat!52}\textbf{100} & \cellcolor{S2Heat!52}\textbf{100} & \cellcolor{S2Heat!52}\textbf{100} & \cellcolor{S2Heat!50}\textbf{98} & \cellcolor{S2Heat!47}\textbf{96} & \cellcolor{S2Heat!44}\textbf{93} & \cellcolor{S2Heat!50}98 & \cellcolor{S2Heat!47}\textbf{96} & \cellcolor{S2Heat!43}\textbf{91} & \cellcolor{S2Heat!45}94 & \cellcolor{S2Heat!36}84 & \cellcolor{S2Heat!33}82 & \cellcolor{S2Heat!28}75 & \cellcolor{S2Heat!51}\textbf{99} & \cellcolor{S2Heat!42}\textbf{91} & \cellcolor{S2Heat!18}62 & \cellcolor{S2Heat!52}\textbf{100} & \cellcolor{S2Heat!51}\textbf{99} & \cellcolor{S2Heat!52}\textbf{100} & \cellcolor{S2Heat!51}\textbf{99} & \cellcolor{S2Heat!39}\textbf{88} & \cellcolor{S2Heat!21}66 & \cellcolor{S2Heat!50}98 & \cellcolor{S2Heat!29}76 & \cellcolor{S2Heat!13}\textbf{53} \\
\cellcolor{S2Heat!15}GPT-6 Luna & \cellcolor{S2Heat!15}79.7 & \cellcolor{S2Heat!36}85 & \cellcolor{S2Heat!51}99 & \cellcolor{S2Heat!30}77 & \cellcolor{S2Heat!33}81 & \cellcolor{S2Heat!52}\textbf{100} & \cellcolor{S2Heat!52}\textbf{100} & \cellcolor{S2Heat!52}\textbf{100} & \cellcolor{S2Heat!48}97 & \cellcolor{S2Heat!50}\textbf{98} & \cellcolor{S2Heat!43}91 & \cellcolor{S2Heat!37}86 & \cellcolor{S2Heat!50}98 & \cellcolor{S2Heat!47}\textbf{96} & \cellcolor{S2Heat!39}88 & \cellcolor{S2Heat!41}90 & \cellcolor{S2Heat!34}82 & \cellcolor{S2Heat!32}80 & \cellcolor{S2Heat!24}70 & \cellcolor{S2Heat!50}98 & \cellcolor{S2Heat!42}\textbf{91} & \cellcolor{S2Heat!18}62 & \cellcolor{S2Heat!52}\textbf{100} & \cellcolor{S2Heat!49}98 & \cellcolor{S2Heat!52}\textbf{100} & \cellcolor{S2Heat!48}96 & \cellcolor{S2Heat!37}86 & \cellcolor{S2Heat!22}\textbf{68} & \cellcolor{S2Heat!51}99 & \cellcolor{S2Heat!29}77 & \cellcolor{S2Heat!10}48 \\
Jev API 1.14 & 60.4 & \cellcolor{S1Heat!37}86 & \cellcolor{S1Heat!51}99 & \cellcolor{S1Heat!30}78 & \cellcolor{S1Heat!15}56 & \cellcolor{S1Heat!52}\textbf{100} & \cellcolor{S1Heat!35}84 & \cellcolor{S1Heat!19}63 & \cellcolor{S1Heat!4}30 & \cellcolor{S1Heat!35}83 & \cellcolor{S1Heat!24}70 & \cellcolor{S1Heat!16}59 & \cellcolor{S1Heat!48}96 & \cellcolor{S1Heat!46}94 & \cellcolor{S1Heat!25}71 & \cellcolor{S1Heat!44}92 & \cellcolor{S1Heat!35}83 & \cellcolor{S1Heat!26}73 & \cellcolor{S1Heat!23}69 & \cellcolor{S1Heat!51}\textbf{99} & \cellcolor{S1Heat!35}83 & \cellcolor{S1Heat!17}60 & \cellcolor{S1Heat!39}88 & \cellcolor{S1Heat!39}88 & \cellcolor{S1Heat!2}20 & \cellcolor{S1Heat!49}97 & \cellcolor{S1Heat!27}75 & \cellcolor{S1Heat!16}59 & \cellcolor{S1Heat!51}99 & \cellcolor{S1Heat!26}73 & \cellcolor{S1Heat!7}41 \\
OpenAI Decisions & 53.6 & \cellcolor{S1Heat!27}74 & \cellcolor{S1Heat!51}99 & \cellcolor{S1Heat!30}77 & \cellcolor{S1Heat!15}58 & \cellcolor{S1Heat!27}74 & \cellcolor{S1Heat!9}46 & \cellcolor{S1Heat!5}34 & \cellcolor{S1Heat!4}30 & \cellcolor{S1Heat!40}88 & \cellcolor{S1Heat!25}71 & \cellcolor{S1Heat!21}66 & \cellcolor{S1Heat!45}94 & \cellcolor{S1Heat!33}81 & \cellcolor{S1Heat!10}46 & \cellcolor{S1Heat!37}86 & \cellcolor{S1Heat!25}71 & \cellcolor{S1Heat!22}67 & \cellcolor{S1Heat!19}63 & \cellcolor{S1Heat!50}98 & \cellcolor{S1Heat!36}84 & \cellcolor{S1Heat!14}55 & \cellcolor{S1Heat!31}79 & \cellcolor{S1Heat!28}75 & \cellcolor{S1Heat!7}40 & \cellcolor{S1Heat!37}86 & \cellcolor{S1Heat!17}61 & \cellcolor{S1Heat!6}38 & \cellcolor{S1Heat!50}98 & \cellcolor{S1Heat!22}68 & \cellcolor{S1Heat!4}31 \\
\midrule
\multicolumn{32}{@{}l}{\textit{Open-weight models: 31B}} \\
KnowLine-31B & 70.2 & \cellcolor{S1Heat!39}88 & \cellcolor{S1Heat!52}\textbf{100} & \cellcolor{S1Heat!35}83 & \cellcolor{S1Heat!30}78 & \cellcolor{S1Heat!50}98 & \cellcolor{S1Heat!50}98 & \cellcolor{S1Heat!25}71 & \cellcolor{S1Heat!7}40 & \cellcolor{S1Heat!43}92 & \cellcolor{S1Heat!31}79 & \cellcolor{S1Heat!22}67 & \cellcolor{S1Heat!50}98 & \cellcolor{S1Heat!46}94 & \cellcolor{S1Heat!13}52 & \cellcolor{S1Heat!44}93 & \cellcolor{S1Heat!38}86 & \cellcolor{S1Heat!35}83 & \cellcolor{S1Heat!29}77 & \cellcolor{S1Heat!51}\textbf{99} & \cellcolor{S1Heat!40}89 & \cellcolor{S1Heat!19}\textbf{64} & \cellcolor{S1Heat!48}97 & \cellcolor{S1Heat!47}96 & \cellcolor{S1Heat!41}90 & \cellcolor{S1Heat!49}97 & \cellcolor{S1Heat!35}84 & \cellcolor{S1Heat!14}56 & \cellcolor{S1Heat!52}\textbf{100} & \cellcolor{S1Heat!27}74 & \cellcolor{S1Heat!9}44 \\
Sieve-31B-v3 & 67.7 & \cellcolor{S1Heat!39}88 & \cellcolor{S1Heat!51}99 & \cellcolor{S1Heat!33}81 & \cellcolor{S1Heat!22}68 & \cellcolor{S1Heat!52}\textbf{100} & \cellcolor{S1Heat!45}94 & \cellcolor{S1Heat!21}66 & \cellcolor{S1Heat!10}47 & \cellcolor{S1Heat!45}93 & \cellcolor{S1Heat!29}77 & \cellcolor{S1Heat!21}66 & \cellcolor{S1Heat!50}98 & \cellcolor{S1Heat!44}93 & \cellcolor{S1Heat!14}55 & \cellcolor{S1Heat!44}93 & \cellcolor{S1Heat!37}86 & \cellcolor{S1Heat!34}82 & \cellcolor{S1Heat!31}\textbf{78} & \cellcolor{S1Heat!50}98 & \cellcolor{S1Heat!36}84 & \cellcolor{S1Heat!19}64 & \cellcolor{S1Heat!48}97 & \cellcolor{S1Heat!49}98 & \cellcolor{S1Heat!24}70 & \cellcolor{S1Heat!48}96 & \cellcolor{S1Heat!37}86 & \cellcolor{S1Heat!17}60 & \cellcolor{S1Heat!51}99 & \cellcolor{S1Heat!17}60 & \cellcolor{S1Heat!8}43 \\
\midrule
\multicolumn{32}{@{}l}{\textit{Open-weight models: 27B}} \\
GLiDE & 70.2 & \cellcolor{S1Heat!41}\textbf{90} & \cellcolor{S1Heat!52}\textbf{100} & \cellcolor{S1Heat!35}\textbf{84} & \cellcolor{S1Heat!28}76 & \cellcolor{S1Heat!50}98 & \cellcolor{S1Heat!37}86 & \cellcolor{S1Heat!20}64 & \cellcolor{S1Heat!4}30 & \cellcolor{S1Heat!45}93 & \cellcolor{S1Heat!23}69 & \cellcolor{S1Heat!15}57 & \cellcolor{S1Heat!48}96 & \cellcolor{S1Heat!38}87 & \cellcolor{S1Heat!14}55 & \cellcolor{S1Heat!46}95 & \cellcolor{S1Heat!38}\textbf{87} & \cellcolor{S1Heat!35}\textbf{84} & \cellcolor{S1Heat!27}75 & \cellcolor{S1Heat!51}\textbf{99} & \cellcolor{S1Heat!36}84 & \cellcolor{S1Heat!15}57 & \cellcolor{S1Heat!52}\textbf{100} & \cellcolor{S1Heat!51}\textbf{99} & \cellcolor{S1Heat!52}\textbf{100} & \cellcolor{S1Heat!49}97 & \cellcolor{S1Heat!33}81 & \cellcolor{S1Heat!20}65 & \cellcolor{S1Heat!51}99 & \cellcolor{S1Heat!33}\textbf{81} & \cellcolor{S1Heat!11}49 \\
Decider-27B & 66.7 & \cellcolor{S1Heat!40}89 & \cellcolor{S1Heat!50}98 & \cellcolor{S1Heat!35}83 & \cellcolor{S1Heat!24}70 & \cellcolor{S1Heat!48}96 & \cellcolor{S1Heat!41}90 & \cellcolor{S1Heat!18}61 & \cellcolor{S1Heat!5}33 & \cellcolor{S1Heat!38}87 & \cellcolor{S1Heat!26}73 & \cellcolor{S1Heat!18}61 & \cellcolor{S1Heat!52}\textbf{100} & \cellcolor{S1Heat!41}90 & \cellcolor{S1Heat!13}52 & \cellcolor{S1Heat!46}95 & \cellcolor{S1Heat!37}86 & \cellcolor{S1Heat!32}81 & \cellcolor{S1Heat!27}75 & \cellcolor{S1Heat!51}\textbf{99} & \cellcolor{S1Heat!39}88 & \cellcolor{S1Heat!15}56 & \cellcolor{S1Heat!51}99 & \cellcolor{S1Heat!48}97 & \cellcolor{S1Heat!20}65 & \cellcolor{S1Heat!49}97 & \cellcolor{S1Heat!35}84 & \cellcolor{S1Heat!22}67 & \cellcolor{S1Heat!51}99 & \cellcolor{S1Heat!31}79 & \cellcolor{S1Heat!11}50 \\
Kev-27B & 61.4 & \cellcolor{S1Heat!36}84 & \cellcolor{S1Heat!49}97 & \cellcolor{S1Heat!33}82 & \cellcolor{S1Heat!22}68 & \cellcolor{S1Heat!52}\textbf{100} & \cellcolor{S1Heat!30}78 & \cellcolor{S1Heat!14}56 & \cellcolor{S1Heat!4}30 & \cellcolor{S1Heat!38}87 & \cellcolor{S1Heat!25}71 & \cellcolor{S1Heat!14}54 & \cellcolor{S1Heat!50}98 & \cellcolor{S1Heat!38}87 & \cellcolor{S1Heat!13}54 & \cellcolor{S1Heat!43}92 & \cellcolor{S1Heat!31}79 & \cellcolor{S1Heat!26}73 & \cellcolor{S1Heat!19}64 & \cellcolor{S1Heat!50}98 & \cellcolor{S1Heat!38}87 & \cellcolor{S1Heat!16}59 & \cellcolor{S1Heat!40}89 & \cellcolor{S1Heat!40}89 & \cellcolor{S1Heat!9}45 & \cellcolor{S1Heat!49}97 & \cellcolor{S1Heat!33}81 & \cellcolor{S1Heat!18}62 & \cellcolor{S1Heat!52}\textbf{100} & \cellcolor{S1Heat!30}78 & \cellcolor{S1Heat!9}45 \\
JevAny-27B & 58.4 & \cellcolor{S1Heat!36}84 & \cellcolor{S1Heat!50}98 & \cellcolor{S1Heat!33}81 & \cellcolor{S1Heat!12}52 & \cellcolor{S1Heat!52}\textbf{100} & \cellcolor{S1Heat!34}82 & \cellcolor{S1Heat!13}53 & \cellcolor{S1Heat!3}27 & \cellcolor{S1Heat!38}87 & \cellcolor{S1Heat!26}73 & \cellcolor{S1Heat!11}50 & \cellcolor{S1Heat!52}\textbf{100} & \cellcolor{S1Heat!38}87 & \cellcolor{S1Heat!17}60 & \cellcolor{S1Heat!46}94 & \cellcolor{S1Heat!33}81 & \cellcolor{S1Heat!27}75 & \cellcolor{S1Heat!20}65 & \cellcolor{S1Heat!43}92 & \cellcolor{S1Heat!24}70 & \cellcolor{S1Heat!15}57 & \cellcolor{S1Heat!45}93 & \cellcolor{S1Heat!41}90 & \cellcolor{S1Heat!5}35 & \cellcolor{S1Heat!50}98 & \cellcolor{S1Heat!35}84 & \cellcolor{S1Heat!16}58 & \cellcolor{S1Heat!51}99 & \cellcolor{S1Heat!24}71 & \cellcolor{S1Heat!7}41 \\
Clef-27B & 56.9 & \cellcolor{S1Heat!36}85 & \cellcolor{S1Heat!49}97 & \cellcolor{S1Heat!31}79 & \cellcolor{S1Heat!19}64 & \cellcolor{S1Heat!48}96 & \cellcolor{S1Heat!27}74 & \cellcolor{S1Heat!10}47 & \cellcolor{S1Heat!1}17 & \cellcolor{S1Heat!43}92 & \cellcolor{S1Heat!24}70 & \cellcolor{S1Heat!17}60 & \cellcolor{S1Heat!50}98 & \cellcolor{S1Heat!43}91 & \cellcolor{S1Heat!12}51 & \cellcolor{S1Heat!46}94 & \cellcolor{S1Heat!30}78 & \cellcolor{S1Heat!27}75 & \cellcolor{S1Heat!17}59 & \cellcolor{S1Heat!49}97 & \cellcolor{S1Heat!39}88 & \cellcolor{S1Heat!18}61 & \cellcolor{S1Heat!39}88 & \cellcolor{S1Heat!35}83 & \cellcolor{S1Heat!1}15 & \cellcolor{S1Heat!43}92 & \cellcolor{S1Heat!37}86 & \cellcolor{S1Heat!14}55 & \cellcolor{S1Heat!50}98 & \cellcolor{S1Heat!26}73 & \cellcolor{S1Heat!8}43 \\
\midrule
\multicolumn{32}{@{}l}{\textit{Open-weight models: 9B}} \\
Clef-Flash-9B & 51.6 & \cellcolor{S1Heat!35}83 & \cellcolor{S1Heat!51}99 & \cellcolor{S1Heat!8}42 & \cellcolor{S1Heat!20}64 & \cellcolor{S1Heat!45}94 & \cellcolor{S1Heat!22}68 & \cellcolor{S1Heat!12}51 & \cellcolor{S1Heat!1}17 & \cellcolor{S1Heat!24}70 & \cellcolor{S1Heat!16}59 & \cellcolor{S1Heat!7}41 & \cellcolor{S1Heat!50}98 & \cellcolor{S1Heat!46}94 & \cellcolor{S1Heat!10}48 & \cellcolor{S1Heat!47}\textbf{96} & \cellcolor{S1Heat!24}70 & \cellcolor{S1Heat!21}66 & \cellcolor{S1Heat!9}46 & \cellcolor{S1Heat!51}\textbf{99} & \cellcolor{S1Heat!35}83 & \cellcolor{S1Heat!15}58 & \cellcolor{S1Heat!28}76 & \cellcolor{S1Heat!16}58 & \cellcolor{S1Heat!5}35 & \cellcolor{S1Heat!42}91 & \cellcolor{S1Heat!11}50 & \cellcolor{S1Heat!4}32 & \cellcolor{S1Heat!41}90 & \cellcolor{S1Heat!14}55 & \cellcolor{S1Heat!3}25 \\
\midrule
\multicolumn{32}{@{}l}{\textit{Open-weight models: 4B}} \\
StartLux Instant & 59.0 & \cellcolor{S1Heat!39}88 & \cellcolor{S1Heat!45}94 & \cellcolor{S1Heat!24}71 & \cellcolor{S1Heat!24}70 & \cellcolor{S1Heat!50}98 & \cellcolor{S1Heat!39}88 & \cellcolor{S1Heat!8}43 & \cellcolor{S1Heat!7}40 & \cellcolor{S1Heat!32}80 & \cellcolor{S1Heat!23}69 & \cellcolor{S1Heat!14}56 & \cellcolor{S1Heat!48}96 & \cellcolor{S1Heat!36}84 & \cellcolor{S1Heat!10}46 & \cellcolor{S1Heat!40}88 & \cellcolor{S1Heat!34}82 & \cellcolor{S1Heat!26}73 & \cellcolor{S1Heat!19}63 & \cellcolor{S1Heat!51}\textbf{99} & \cellcolor{S1Heat!34}82 & \cellcolor{S1Heat!14}54 & \cellcolor{S1Heat!37}86 & \cellcolor{S1Heat!32}80 & \cellcolor{S1Heat!14}55 & \cellcolor{S1Heat!48}96 & \cellcolor{S1Heat!31}79 & \cellcolor{S1Heat!13}54 & \cellcolor{S1Heat!50}98 & \cellcolor{S1Heat!6}39 & \cellcolor{S1Heat!4}31 \\
StartLux Auto & 58.6 & \cellcolor{S1Heat!35}84 & \cellcolor{S1Heat!48}96 & \cellcolor{S1Heat!24}71 & \cellcolor{S1Heat!24}70 & \cellcolor{S1Heat!50}98 & \cellcolor{S1Heat!41}90 & \cellcolor{S1Heat!33}81 & \cellcolor{S1Heat!10}47 & \cellcolor{S1Heat!32}80 & \cellcolor{S1Heat!19}63 & \cellcolor{S1Heat!14}54 & \cellcolor{S1Heat!41}90 & \cellcolor{S1Heat!37}86 & \cellcolor{S1Heat!12}51 & \cellcolor{S1Heat!36}85 & \cellcolor{S1Heat!27}74 & \cellcolor{S1Heat!21}66 & \cellcolor{S1Heat!18}62 & \cellcolor{S1Heat!50}98 & \cellcolor{S1Heat!34}82 & \cellcolor{S1Heat!15}56 & \cellcolor{S1Heat!39}88 & \cellcolor{S1Heat!29}76 & \cellcolor{S1Heat!5}35 & \cellcolor{S1Heat!43}92 & \cellcolor{S1Heat!29}76 & \cellcolor{S1Heat!9}46 & \cellcolor{S1Heat!43}92 & \cellcolor{S1Heat!11}50 & \cellcolor{S1Heat!4}31 \\
H2O Lightning & 54.9 & \cellcolor{S1Heat!37}85 & \cellcolor{S1Heat!45}94 & \cellcolor{S1Heat!21}66 & \cellcolor{S1Heat!19}63 & \cellcolor{S1Heat!50}98 & \cellcolor{S1Heat!25}72 & \cellcolor{S1Heat!9}44 & \cellcolor{S1Heat!8}43 & \cellcolor{S1Heat!24}70 & \cellcolor{S1Heat!14}54 & \cellcolor{S1Heat!9}46 & \cellcolor{S1Heat!43}92 & \cellcolor{S1Heat!32}80 & \cellcolor{S1Heat!8}44 & \cellcolor{S1Heat!42}91 & \cellcolor{S1Heat!34}82 & \cellcolor{S1Heat!22}68 & \cellcolor{S1Heat!19}64 & \cellcolor{S1Heat!50}98 & \cellcolor{S1Heat!38}87 & \cellcolor{S1Heat!18}61 & \cellcolor{S1Heat!36}85 & \cellcolor{S1Heat!32}80 & \cellcolor{S1Heat!7}40 & \cellcolor{S1Heat!48}96 & \cellcolor{S1Heat!21}66 & \cellcolor{S1Heat!6}38 & \cellcolor{S1Heat!45}94 & \cellcolor{S1Heat!6}36 & \cellcolor{S1Heat!3}28 \\
Decider Flash & 52.0 & \cellcolor{S1Heat!39}88 & \cellcolor{S1Heat!50}98 & \cellcolor{S1Heat!27}75 & \cellcolor{S1Heat!17}59 & \cellcolor{S1Heat!48}96 & \cellcolor{S1Heat!28}76 & \cellcolor{S1Heat!7}40 & \cellcolor{S1Heat!2}20 & \cellcolor{S1Heat!29}77 & \cellcolor{S1Heat!20}64 & \cellcolor{S1Heat!13}53 & \cellcolor{S1Heat!43}92 & \cellcolor{S1Heat!34}83 & \cellcolor{S1Heat!7}40 & \cellcolor{S1Heat!42}91 & \cellcolor{S1Heat!31}79 & \cellcolor{S1Heat!20}65 & \cellcolor{S1Heat!17}61 & \cellcolor{S1Heat!51}\textbf{99} & \cellcolor{S1Heat!32}80 & \cellcolor{S1Heat!15}56 & \cellcolor{S1Heat!39}88 & \cellcolor{S1Heat!35}83 & \cellcolor{S1Heat!1}15 & \cellcolor{S1Heat!46}95 & \cellcolor{S1Heat!22}68 & \cellcolor{S1Heat!7}41 & \cellcolor{S1Heat!45}94 & \cellcolor{S1Heat!14}56 & \cellcolor{S1Heat!5}34 \\
Jeva-4B & 51.6 & \cellcolor{S1Heat!33}81 & \cellcolor{S1Heat!38}87 & \cellcolor{S1Heat!18}61 & \cellcolor{S1Heat!18}62 & \cellcolor{S1Heat!35}84 & \cellcolor{S1Heat!18}62 & \cellcolor{S1Heat!13}53 & \cellcolor{S1Heat!2}23 & \cellcolor{S1Heat!33}82 & \cellcolor{S1Heat!20}64 & \cellcolor{S1Heat!11}50 & \cellcolor{S1Heat!34}82 & \cellcolor{S1Heat!26}73 & \cellcolor{S1Heat!17}60 & \cellcolor{S1Heat!28}76 & \cellcolor{S1Heat!22}68 & \cellcolor{S1Heat!20}64 & \cellcolor{S1Heat!12}51 & \cellcolor{S1Heat!48}96 & \cellcolor{S1Heat!36}85 & \cellcolor{S1Heat!19}\textbf{64} & \cellcolor{S1Heat!36}85 & \cellcolor{S1Heat!31}79 & \cellcolor{S1Heat!9}45 & \cellcolor{S1Heat!26}73 & \cellcolor{S1Heat!4}32 & \cellcolor{S1Heat!1}17 & \cellcolor{S1Heat!36}85 & \cellcolor{S1Heat!18}61 & \cellcolor{S1Heat!1}18 \\
JevAny-4B & 50.9 & \cellcolor{S1Heat!35}83 & \cellcolor{S1Heat!49}97 & \cellcolor{S1Heat!22}67 & \cellcolor{S1Heat!17}61 & \cellcolor{S1Heat!43}92 & \cellcolor{S1Heat!17}60 & \cellcolor{S1Heat!6}39 & \cellcolor{S1Heat!2}20 & \cellcolor{S1Heat!26}73 & \cellcolor{S1Heat!21}66 & \cellcolor{S1Heat!10}47 & \cellcolor{S1Heat!48}96 & \cellcolor{S1Heat!36}84 & \cellcolor{S1Heat!8}44 & \cellcolor{S1Heat!41}90 & \cellcolor{S1Heat!23}69 & \cellcolor{S1Heat!19}63 & \cellcolor{S1Heat!9}46 & \cellcolor{S1Heat!42}91 & \cellcolor{S1Heat!25}72 & \cellcolor{S1Heat!12}52 & \cellcolor{S1Heat!25}71 & \cellcolor{S1Heat!19}63 & \cellcolor{S1Heat!9}45 & \cellcolor{S1Heat!44}93 & \cellcolor{S1Heat!25}72 & \cellcolor{S1Heat!7}41 & \cellcolor{S1Heat!45}94 & \cellcolor{S1Heat!10}48 & \cellcolor{S1Heat!1}19 \\
SII-4B & 48.0 & \cellcolor{S1Heat!30}78 & \cellcolor{S1Heat!42}91 & \cellcolor{S1Heat!19}64 & \cellcolor{S1Heat!7}40 & \cellcolor{S1Heat!18}62 & \cellcolor{S1Heat!8}42 & \cellcolor{S1Heat!6}39 & \cellcolor{S1Heat!3}27 & \cellcolor{S1Heat!24}70 & \cellcolor{S1Heat!16}59 & \cellcolor{S1Heat!12}51 & \cellcolor{S1Heat!41}90 & \cellcolor{S1Heat!33}81 & \cellcolor{S1Heat!9}45 & \cellcolor{S1Heat!35}84 & \cellcolor{S1Heat!25}71 & \cellcolor{S1Heat!17}60 & \cellcolor{S1Heat!9}45 & \cellcolor{S1Heat!46}95 & \cellcolor{S1Heat!36}84 & \cellcolor{S1Heat!14}54 & \cellcolor{S1Heat!34}82 & \cellcolor{S1Heat!24}71 & \cellcolor{S1Heat!17}60 & \cellcolor{S1Heat!35}84 & \cellcolor{S1Heat!8}43 & \cellcolor{S1Heat!2}20 & \cellcolor{S1Heat!42}91 & \cellcolor{S1Heat!11}49 & \cellcolor{S1Heat!1}15 \\
Kev-4B & 46.1 & \cellcolor{S1Heat!27}74 & \cellcolor{S1Heat!43}92 & \cellcolor{S1Heat!19}64 & \cellcolor{S1Heat!11}50 & \cellcolor{S1Heat!41}90 & \cellcolor{S1Heat!16}58 & \cellcolor{S1Heat!4}33 & \cellcolor{S1Heat!2}20 & \cellcolor{S1Heat!21}67 & \cellcolor{S1Heat!21}66 & \cellcolor{S1Heat!14}56 & \cellcolor{S1Heat!39}88 & \cellcolor{S1Heat!27}74 & \cellcolor{S1Heat!7}41 & \cellcolor{S1Heat!34}82 & \cellcolor{S1Heat!21}66 & \cellcolor{S1Heat!15}58 & \cellcolor{S1Heat!4}32 & \cellcolor{S1Heat!44}93 & \cellcolor{S1Heat!32}80 & \cellcolor{S1Heat!13}54 & \cellcolor{S1Heat!28}76 & \cellcolor{S1Heat!25}72 & \cellcolor{S1Heat!11}50 & \cellcolor{S1Heat!33}81 & \cellcolor{S1Heat!9}46 & \cellcolor{S1Heat!2}22 & \cellcolor{S1Heat!35}83 & \cellcolor{S1Heat!7}39 & \cellcolor{S1Heat!1}16 \\
\midrule
\multicolumn{32}{@{}l}{\textit{Open-weight models: 0.4B}} \\
Laya & 30.9 & \cellcolor{S1Heat!20}65 & \cellcolor{S1Heat!34}82 & \cellcolor{S1Heat!0}7 & \cellcolor{S1Heat!4}30 & \cellcolor{S1Heat!8}42 & \cellcolor{S1Heat!8}42 & \cellcolor{S1Heat!2}23 & \cellcolor{S1Heat!2}20 & \cellcolor{S1Heat!16}58 & \cellcolor{S1Heat!12}51 & \cellcolor{S1Heat!12}51 & \cellcolor{S1Heat!3}26 & \cellcolor{S1Heat!3}29 & \cellcolor{S1Heat!2}24 & \cellcolor{S1Heat!8}43 & \cellcolor{S1Heat!11}49 & \cellcolor{S1Heat!7}39 & \cellcolor{S1Heat!5}35 & \cellcolor{S1Heat!37}86 & \cellcolor{S1Heat!18}62 & \cellcolor{S1Heat!6}38 & \cellcolor{S1Heat!12}52 & \cellcolor{S1Heat!8}44 & \cellcolor{S1Heat!4}30 & \cellcolor{S1Heat!0}1 & \cellcolor{S1Heat!0}1 & \cellcolor{S1Heat!0}0 & \cellcolor{S1Heat!9}44 & \cellcolor{S1Heat!0}6 & \cellcolor{S1Heat!0}2 \\
\bottomrule
\end{tabular}

\end{table}

\subsubsection{Reliability}
\label{sec:exp-reliability}
\begin{figure}[t]
  \centering
  \includegraphics[width=\linewidth]{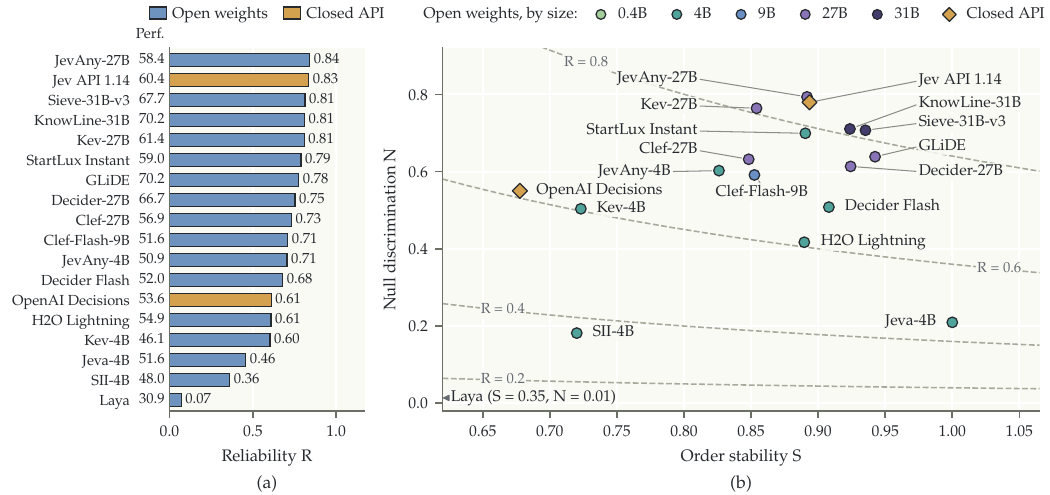}
  \captionsetup{font=footnotesize}
  \caption{\textbf{Reliability and its two components} for the 18 System One models. (a) Reliability $R$ (\Cref{eq:reliability-score}), with each model's performance (Perf.), its leaderboard score (\Cref{eq:capability-score}), to the left of its bar. (b) $N$ against $S$; 0 is chance level for both, and dashed curves mark constant $R$.}
  \label{fig:reliability}
\end{figure}

We measure reliability $R$ (\Cref{eq:reliability-score}) and its components, order stability $S$ (\Cref{eq:order-stability}) and null discrimination $N$ (\Cref{eq:null-discrimination}), for the 18 System One models (\Cref{fig:reliability}). While performance and reliability are positively correlated sometimes, higher performance does not always imply higher reliability: the two most reliable models, JevAny-27B and Jev API 1.14, rank only eighth and sixth of the 18 in performance (\Cref{tab:tier-scores}).

\paragraph{Option order stability.}
Most models keep their answer when only the order of the options changes: 14 of the 18 reach $S>0.8$ (\Cref{fig:reliability}b). Apart from Jeva-4B, at $S=1$ by construction (\Cref{app:configs}), these 14 lie between 0.83 and 0.94, while the remaining four, Laya (0.35), SII-4B, Kev-4B, and OpenAI Decisions (0.68 to 0.72), are sensitive to the order.

\paragraph{Null discrimination.}
Null discrimination, in contrast, varies widely, from near chance to about 0.8 (\Cref{fig:reliability}b). Size alone does not decide it: within each family released in two sizes the larger model does score higher on null discrimination, yet some smaller 4B models exceed some larger 9B and 27B models.

\subsection{Discussion of Results}
\label{sec:discussion}
The main results raise two questions. Since System One models trail the System Two references on general performance, \Cref{sec:analysis-domains} asks in which domains the gap lies and whether a more capable backbone or reasoning closes it. \Cref{sec:analysis-reliability} asks how far performance explains each reliability component.

\subsubsection{Where System One Models Match System Two, and What Closes the Gap}
\label{sec:analysis-domains}

\begin{figure}[t]
  \centering
  \includegraphics[width=\linewidth]{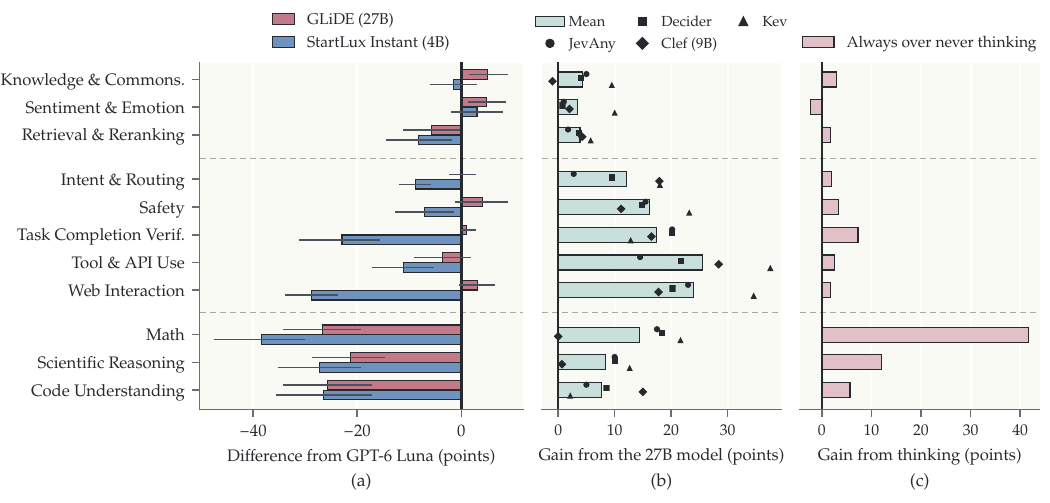}
  \captionsetup{font=footnotesize}
  \caption{\textbf{The gap to a System Two model, and what narrows it, differ by domain.} (a) Score difference of two System One models from GPT-6 Luna on the medium and hard tiers, with 95\% confidence intervals (\Cref{app:analysis}). (b) Gain from the smaller to the 27B model of each family. (c) Gain of StartLux Auto from always thinking over never thinking. Rows are grouped by the three categories of \Cref{sec:analysis-domains}.}
  \label{fig:analysis-domains}
\end{figure}

\paragraph{The gap to System Two models across domains.}
How close a System One model's performance comes to a System Two model depends on the domain. We compare GPT-6 Luna with the best-performing open-weight System One models of two sizes: StartLux Instant (4B) and GLiDE (27B) (\Cref{fig:analysis-domains}a). By their nature and by how the models perform on them, the domains fall into three categories: direct semantic judgment, where System One models perform comparably to System Two or even better; fine-grained grounded decision, where only the System One model with a larger, more capable backbone comes close to System Two; and multi-step derivation, where both System One models fall far behind.

To see what drives the gap in each category, we measure two gains (\Cref{fig:analysis-domains}b and c). The gain from backbone capability compares the smaller and the larger model of each family released in both sizes: every smaller model builds on Qwen3.5, every larger model on the newer Qwen3.8-27B. The gain from thinking compares StartLux Auto with thinking always on against never on (\Cref{app:analysis}).

\paragraph{Direct semantic judgment.}
In Knowledge \& Commonsense, Sentiment \& Emotion Understanding, and Retrieval \& Reranking, System One models perform comparably to Luna (\Cref{fig:analysis-domains}a). Neither a more capable backbone nor thinking adds much (\Cref{fig:analysis-domains}b and c).

\paragraph{Fine-grained grounded decision.}
In Intent \& Routing, Safety, Task Completion Verification, Tool \& API Use, and Web Interaction, GLiDE comes within 4 points of Luna, while StartLux Instant trails by up to 29 (\Cref{fig:analysis-domains}a). The model with larger and newer backbone of every family scores higher in all five domains, by far more than thinking adds (\Cref{fig:analysis-domains}b and c).

\paragraph{Multi-step derivation.}
In Math, Scientific Reasoning, and Code Understanding, both trail Luna significantly (\Cref{fig:analysis-domains}a). Thinking helps most in Math and Scientific Reasoning (\Cref{fig:analysis-domains}c): in Math it lifts StartLux Auto from 50.8 to 92.5, close to Luna, in line with the finding that chain-of-thought helps mainly on math and symbolic reasoning~\citep{sprague2025cot}.

\subsubsection{Factors Behind Reliability}
\label{sec:analysis-reliability}

\begin{figure}[t]
  \centering
  \includegraphics[width=\linewidth]{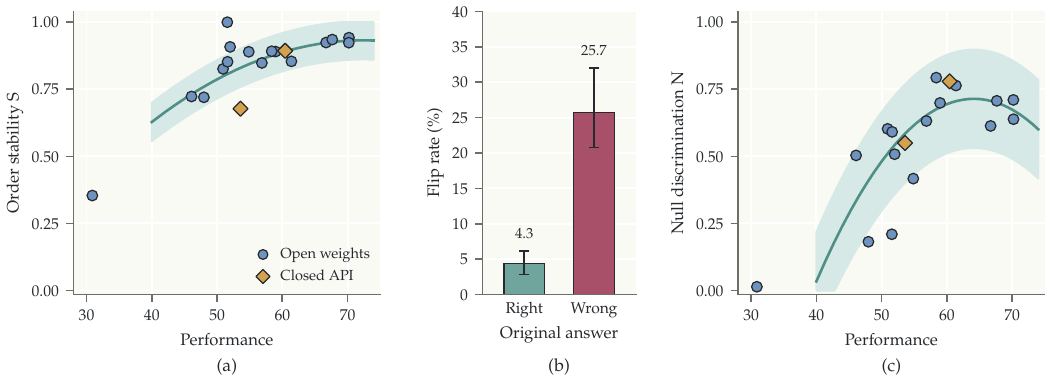}
  \captionsetup{font=footnotesize}
  \caption{\textbf{Order stability rises with performance; null discrimination follows it only loosely.} In (a) and (c), each point is one of the 18 System One models, the line is a quadratic trend, and the band holds 90\% of the fitted models (\Cref{app:analysis}). (a) Order stability $S$ (\Cref{eq:order-stability}) against performance. (b) Share of option reorderings that change the answer, for originally right and wrong answers: median and interquartile range over the models. (c) Null discrimination $N$ (\Cref{eq:null-discrimination}) against performance.}
  \label{fig:analysis-reliability}
\end{figure}

\paragraph{Option order stability.}
Order stability largely follows performance (\Cref{fig:analysis-reliability}a): it rises with performance, and most models lie close to the trend. The flip rates further demonstrate this: the model changes its answer much more frequently on the reorderings of questions it answers wrongly, against those it answers correctly (\Cref{fig:analysis-reliability}b). Instability thus arises mostly where a model performs bad.

\paragraph{Null discrimination.}
Null discrimination ($N$) follows performance only loosely (\Cref{fig:analysis-reliability}c): around the same performance, it varies far more than option order stability. Kev-27B, Jev API 1.14, and JevAny-27B each exceed the four best-performing System One models in $N$ (\Cref{app:analysis}). A performance score therefore says much about a model's order stability but little about whether it notices a missing answer, which has to be measured directly.

\section{Related Work}
\label{sec:related_work}

\textbf{System One Models and Decision Benchmarks.}
System One models expose predictions over explicitly specified outcomes, enabling classification, routing, scoring, and verification without open-ended text generation. Jev introduced this interface as a general-purpose service~\citep{almeida2026jev}, while open releases such as Kev and JevAny provide language-model-based implementations~\citep{palmer2026kev4b,wei2026jevany}. Recent research investigates how these capabilities can be obtained from pretrained models: SII-Decision distills reasoning-informed predictions into a single-pass readout, whereas Releasing the Inner Jev studies answer-token readout and supervised adaptation~\citep{duan2026general,geuter2026innerjev}. Evaluation efforts already extend beyond accuracy. JevBench measures capability, calibration, latency, and cost~\citep{jevbench}; JEVal covers ten application domains and examines how local decisions compose into agent workflows and social simulations~\citep{duan2026general}. OpenDecisions additionally includes difficulty tiers and controlled perturbations~\citep{geuter2026innerjev}. Our work complements these efforts with within-domain difficulty tiers, controlled history-dependent verification, and a separate reliability score within a broader evaluation of decision competence.

\textbf{Generalist Capability Evaluation.}
MMLU measures knowledge and problem solving across academic and professional subjects~\citep{hendrycks2021mmlu}, while BIG-Bench Hard probes challenging tasks and the effects of explicit reasoning~\citep{suzgun2023bbh}. Humanity's Last Exam extends evaluation toward expert-level academic questions~\citep{hle2026benchmark}. These benchmarks establish useful measures of capability breadth and demanding problem solving. HELM further emphasizes standardized evaluation across scenarios and multiple desiderata, including calibration and efficiency~\citep{liang2023holistic}. SO-Arena adopts a complementary focus: assessing generalist competence through decisions over supplied evidence and specified alternatives, while also examining whether judgments remain reliable as their informational and temporal requirements change. Academic reasoning is therefore one component of our evaluation, alongside semantic, policy, evidence, and environment-dependent decisions.

\textbf{Judges, Reward Models, and Verification Reliability.}
Research on LLM-as-a-judge establishes the usefulness of automated evaluation while documenting position, verbosity, and self-enhancement biases~\citep{zheng2023judging}. RewardBench evaluates preferences across chat, safety, and reasoning, whereas JudgeBench constructs response comparisons grounded in factual and logical correctness~\citep{lambert2025rewardbench,tan2025judgebench}. VerifyBench examines agreement between candidate responses and reference answers across scientific domains, including sensitivity to response presentation~\citep{li2026verifybench}. These studies motivate scrutinizing the evaluator itself. Separately, calibration research distinguishes predictive accuracy from the quality of confidence estimates, and selective classification studies the trade-off between error and coverage~\citep{guo2017calibration,geifman2017selective}. We connect these perspectives to general-purpose decision evaluation, encompassing judgments whose evidence consists of documents, structured states, or interaction histories rather than only candidate answer pairs.

\textbf{Evaluation in Agentic Environments.}
Mind2Web evaluates action prediction across tasks, websites, and domains~\citep{deng2023mind2web}. WebArena provides reproducible web environments and functional evaluation using answers and programmatic state checks~\citep{zhou2024webarena}; $\tau$-bench evaluates policy-constrained tool interaction using database outcomes and repeated-trial reliability~\citep{yao2025taubench}. More recently, VerifierBench evaluates the generation of executable checkers against valid solutions, ordinary errors, and adversarial cases~\citep{yin2026verifierbench}. These works address complementary evaluation targets: executing tasks, maintaining reliable behavior, or constructing verification programs. We evaluate decision models in two of these roles: as the policy that chooses each action of a tool or web episode from a menu of public options, and as the verifier that judges recorded executions, including trajectories whose final goal projections agree while their histories differ (\Cref{app:construction-evidence}).

\section{Conclusion}
\label{sec:conclusion}
We introduce \textbf{SO-Arena}, a benchmark assessing generalist competence and decision reliability in System One models across eleven domains. It combines difficulty-stratified tasks, action selection and completion verification, and controlled option-set interventions to evaluate performance beyond aggregate accuracy.
Our evaluation of 21 models reveals substantial capability gaps on demanding tasks, especially in agentic settings, despite the latency advantages of System One models. Stronger backbones and additional reasoning offer domain-dependent gains rather than universal improvements. More importantly, strong competence does not guarantee reliable judgment: option-order stability largely tracks performance, while recognition of missing correct options varies considerably among similarly capable models. These findings highlight the need to advance both capability and reliability in future System One models.

\FloatBarrier
{\small\bibliographystyle{plainnat}\bibliography{lastexam_refs}}

\newpage
\appendix
\crefalias{section}{appendix}
\crefalias{subsection}{appendix}
\raggedbottom
\Needspace{5\baselineskip}
\section{Benchmark Construction Details}
\label{app:construction}

This appendix complements \Cref{sec:bench} with per-domain construction details. Every item records its source, pinned version, original split, and original identifier; these records, together with the reference of each item, are kept separate from the model inputs.

\begin{table}[htp]
  \centering
  \captionsetup{font=small}
  \caption{\textbf{Composition of SO-Arena.} Counts are natural tasks. Tiers give easy/medium/hard counts; Math additionally has 30 very hard AIME problems, and ``--'' marks domains without tiers. Where tiers draw on different sources, sources are listed from easy to hard, separated by $\rightarrow$. ``Seq.''\ denotes a sequence of \texttt{choice} requests.}
  \label{tab:bench-composition}
  \footnotesize
  \setlength{\tabcolsep}{4pt}
  \renewcommand{\arraystretch}{1.05}
  \begin{tabularx}{\linewidth}{@{}>{\raggedright\arraybackslash}p{3.75cm} r c l >{\raggedright\arraybackslash}X@{}}
    \toprule
    \textbf{Domain} & \textbf{Tasks} & \textbf{Tiers} & \textbf{Format} & \textbf{Sources} \\
    \midrule
    \rowcolor{TabGroup}\multicolumn{5}{@{}c@{}}{\textit{Semantic understanding}} \\
    Sentiment \& Emotion Understanding & 200 & -- & \texttt{choice}, \texttt{score} & Amazon Reviews 2023~\citep{hou2024amazonreviews}, GoEmotions~\citep{demszky2020goemotions}, SST-5~\citep{socher2013sst}, DynaSent~\citep{potts2021dynasent}, ARTS~\citep{xing2020arts} \\
    Intent \& Routing & 300 & 100/100/100 & \texttt{choice}, \texttt{noul} & CLINC150~\citep{larson2019clinc}, BANKING77~\citep{casanueva2020banking77}, SGD~\citep{rastogi2020sgd} $\rightarrow$ NatCS~\citep{gung2023natcs} $\rightarrow$ NLU++~\citep{casanueva2022nlupp} \\
    \rowcolor{TabGroup}\multicolumn{5}{@{}c@{}}{\textit{Knowledge and reasoning}} \\
    Math & 200 & 50/50/70 & \texttt{choice} & GSM8K-Platinum~\citep{cobbe2021gsm8k,vendrow2025reliability}, MATH500~\citep{hendrycks2021math,lightman2023verify}, AIME 2025--2026~\citep{balunovic2025matharena} \\
    Code Understanding & 200 & 60/70/70 & \texttt{noul} & CodeJudgeBench~\citep{jiang2025codejudgebench}, LiveCodeBench~\citep{jain2024livecodebench}, LeetCodeDataset~\citep{xia2025leetcodedataset} \\
    Scientific Reasoning & 200 & 50/70/80 & \texttt{choice} & ARC-Challenge~\citep{clark2018arc} $\rightarrow$ MMLU-Redux~\citep{gema2024mmluredux} $\rightarrow$ GPQA Diamond~\citep{rein2023gpqa} \\
    Knowledge \& Commonsense & 200 & -- & \texttt{choice} & MMLU-Redux~\citep{hendrycks2021mmlu,gema2024mmluredux}, HellaSwag~\citep{zellers2019hellaswag}, CommonsenseQA~\citep{talmor2019commonsenseqa} \\
    \rowcolor{TabGroup}\multicolumn{5}{@{}c@{}}{\textit{Safety and policy compliance}} \\
    Safety & 500 & 177/165/158 & \texttt{choice}, \texttt{noul} & RAGTruth~\citep{niu2024ragtruth}, ReasoningShield~\citep{li2025reasoningshield}, HarmThoughts~\citep{kakkar2026harmthoughts}, DynaBench~\citep{hoover2025dynaguard}, Aegis~2.0~\citep{ghosh2025aegis2}, OR-Bench~\citep{cui2024orbench} \\
    \rowcolor{TabGroup}\multicolumn{5}{@{}c@{}}{\textit{Evidence assessment}} \\
    Retrieval \& Reranking & 300 & 100/100/100 & \texttt{noul}, \texttt{score} & TREC DL~\citep{craswell2020trecdl,craswell2021trecdl} $\rightarrow$ BRIGHT~\citep{su2024bright}, Touch\'e 2020~\citep{bondarenko2020overview,thakur2021beir} $\rightarrow$ ACORD~\citep{wang2025acord} \\
    Task Completion Verification & 200 & 91/89/20 & \texttt{choice}, \texttt{noul} & BrowseComp-Plus~\citep{chen2025browsecompplus}, CALVIN~\citep{mees2022calvin}, $\tau$-bench~\citep{yao2025taubench}, MiniGrid~\citep{chevalierboisvert2023minigrid}, EAI-VirtualHome~\citep{li2024embodied,puig2018virtualhome} \\
    \rowcolor{TabGroup}\multicolumn{5}{@{}c@{}}{\textit{Agent task completion}} \\
    Tool \& API Use & 282 & 100/78/104 & live seq. & BFCL~\citep{patil2025bfcl} $\rightarrow$ ToolSandbox~\citep{lu-etal-2025-toolsandbox} $\rightarrow$ BFCL \\
    Text-based Web Interaction & 337 & 100/137/100 & \texttt{choice}, \texttt{noul}; seq. & Mind2Web~\citep{deng2023mind2web} $\rightarrow$ BreakingWeb~\citep{yin2026breakingweb} \\
    \bottomrule
  \end{tabularx}
\end{table}

\subsection{Semantic Understanding}
\label{app:construction-semantic}

\paragraph{Sentiment \& Emotion Understanding (200).}
We construct 200 items from five public datasets. From the All Beauty
category of Amazon Reviews 2023~\citep{hou2024amazonreviews}, we sample
50 reviews, with 10 for each recorded star rating. We sample 40
single-label test comments from GoEmotions~\citep{demszky2020goemotions},
with four per target emotion across ten emotions. For sentiment
classification, we sample 30 SST-5~\citep{socher2013sst} sentences
(six per class) and 50 DynaSent~\citep{potts2021dynasent} test sentences
(18 negative, 18 positive, and 14 neutral). We also construct 30
aspect-level sentiment items from 10 restaurant sentences in
ARTS~\citep{xing2020arts}.
The items use two decision formats. In ordinal \texttt{score} items,
the model reads a review title and text and predicts the reviewer's
rating on the original 1--5 star scale. In single-label classification
items, the model selects one label from a shuffled list of options.
GoEmotions items use the full set of 28 labels, including labels
outside the ten emotions used for sampling. We score predictions
against the recorded star rating or the source label, depending on
the format. The resulting scores reflect agreement with the reviewer
or annotator, since a review may leave the exact star rating ambiguous
and a comment may admit more than one plausible emotion label.

\paragraph{Intent \& Routing (300).}
We construct 300 items, with 100 per difficulty tier. For the easy tier, we sample 42 single requests from CLINC150~\citep{larson2019clinc}, 40 from BANKING77~\citep{casanueva2020banking77}, and 18 from SGD~\citep{rastogi2020sgd}. The BANKING77 sample contains 29 in-scope and 11 out-of-scope requests relative to a 50-category service catalog. For the medium tier, we select 100 complete NatCS~\citep{gung2023natcs} customer-service conversations (36 insurance, 30 banking, and 34 finance). For the hard tier, we select 100 NLU++~\citep{casanueva2022nlupp} utterances (73 banking and 27 hotels). We rank medium and hard candidates using a deterministic hash and screen them using only the source data. The audits set aside 30 medium and 20 hard candidates because of label uncertainty or poor task fit. The tiers differ in what the model must identify. Easy items ask the model to judge the intent of a single request. Medium items require tracking exactly two customer goals across a full conversation. The goals may appear far apart, and the model must distinguish them from goals suggested only by the agent. Hard items require identifying which components of a larger domain ontology apply to an utterance. The easy tier uses 47 four-way \texttt{choice} items and 53 binary \texttt{noul} items. Each \texttt{choice} item presents the native intent and three neighboring native classes, while each \texttt{noul} item asks whether one candidate intent applies. In the medium tier, we pose one \texttt{noul} question per native goal label, yielding 22--39 questions per conversation. In the hard tier, we pose one \texttt{noul} question per ontology component, yielding 48 questions for banking and 40 for hotels. We use the native component definitions verbatim. For medium and hard items, we form the predicted label set from all labels whose affirmative probability exceeds \(0.5\). We score each conversation or utterance using the Jaccard index between this set and the source-annotated label set.

\subsection{Knowledge and Reasoning}
\label{app:construction-knowledge}

\paragraph{Math (200).}
We sample 50 problems from GSM8K-Platinum~\citep{cobbe2021gsm8k,vendrow2025reliability}, 120 from MATH500~\citep{hendrycks2021math,lightman2023verify}, and 30 from AIME 2025 and 2026~\citep{balunovic2025matharena}. We restrict the MATH500 sample to problems with numeric answers and no figures. We assign difficulty using calculation-step counts for GSM8K-Platinum and source levels for MATH500. The easy tier combines 30 GSM8K-Platinum problems requiring at most three steps and 20 MATH500 Level~1 problems. The medium tier combines 20 GSM8K-Platinum problems requiring four or more steps and 30 MATH500 Levels~2--3 problems. The hard tier contains 70 MATH500 Levels~4--5 problems, and the very hard tier contains 13 AIME 2025 and 17 AIME 2026 problems. We convert each problem into a multiple-choice item. Of the 600 distractors, 350 are incorrect answers obtained by sampling Qwen3.5-4B-Base 16 times per problem without reasoning, 22 are intermediate results from GSM8K solutions, and 228 are rule-based near misses. We sort options by numeric value, balance the position of the correct answer, and exclude items with duplicate options, doubtful labels, or a correct answer that can be identified from the options alone. The correct numeric answer serves as the scoring reference.

\paragraph{Code Understanding (200).}
We take 156 solutions written by seven LLMs from CodeJudgeBench~\citep{jiang2025codejudgebench} and sample 44 additional solutions from Qwen3.5-4B-Base for 2024--2025 LeetCodeDataset~\citep{xia2025leetcodedataset} Easy problems outside CodeJudgeBench. We compute a size score as the mean of the standardized log line count and standardized log branch and loop count. For correct solutions we add the LeetCode difficulty, encoded as $-1$, $0$, or $1$. For incorrect solutions we add 1.5 times the standardized fraction of tests passed. Within each label, the bottom quartile forms the easy pool, the second quartile the medium pool, and the top quartile the hard pool. An integer program selects 60, 70, and 70 items from these pools, with equal numbers of correct and incorrect solutions per tier and one item per problem family. It matches the labels within each tier by LeetCode difficulty, generator, and code length. Each item presents a LeetCode problem statement and a Python solution with comments and docstrings removed, and asks whether the solution returns the correct answer for every valid input. We establish the reference label by re-executing each solution under its original import header on the public and private LiveCodeBench~\citep{jain2024livecodebench} tests in a network-isolated sandbox. For solutions labeled correct, we additionally require either a pass on the corresponding LeetCodeDataset tests or a recorded manual check, and compare each solution with an independent brute-force implementation on thousands of valid random inputs. For solutions labeled incorrect, we require a wrong answer or runtime error on a problem example or a test whose expected output is independently confirmed. We exclude solutions whose only failures arise from machine-dependent time, memory, or recursion limits.

\paragraph{Scientific Reasoning (200).}
We sample 50 four-option questions from the ARC-Challenge~\citep{clark2018arc} test set, 70 natural-science questions marked error-free in MMLU-Redux~\citep{gema2024mmluredux}, and 80 GPQA Diamond~\citep{rein2023gpqa} questions (36 chemistry, 35 physics, and 9 biology). We exclude items with duplicate option texts and those flagged as incorrect or ambiguous in an earlier internal label audit. We assign difficulty by source, with ARC-Challenge in the easy tier, MMLU-Redux in the medium tier, and GPQA Diamond in the hard tier. Each item asks the model to select one answer from the available options. We shuffle the GPQA Diamond options with a fixed seed, balance the correct-answer position within each source, and use the audited answer key as the scoring reference.

\paragraph{Knowledge \& Commonsense (200).}
We sample 120 MMLU-Redux~\citep{gema2024mmluredux} questions marked error-free across 36 non-STEM subjects, with at most six per subject, along with 40 HellaSwag~\citep{zellers2019hellaswag} and 40 CommonsenseQA~\citep{talmor2019commonsenseqa} validation items. Within the MMLU-Redux material, we reserve natural-science subjects for Scientific Reasoning and exclude mathematics, computer science, and logic from both domains. We also exclude items with duplicate option texts and those flagged as incorrect or ambiguous in an earlier internal label audit. We leave this domain untiered because its tasks offer limited scope for systematically varying decision complexity. Each item asks the model to select one answer from the available options. For CommonsenseQA, we randomly remove one distractor to leave four options. We balance the correct-answer position within each source and use the audited answer key as the scoring reference.

\subsection{Safety and Policy Compliance}
\label{app:construction-safety}

\paragraph{Safety (500).}
We construct 500 items from six sources, drawing 150 from RAGTruth~\citep{niu2024ragtruth}, 75 from ReasoningShield~\citep{li2025reasoningshield}, 50 from HarmThoughts~\citep{kakkar2026harmthoughts}, 100 from DynaBench~\citep{hoover2025dynaguard}, 100 from Aegis~2.0~\citep{ghosh2025aegis2}, and 25 from the hard-benign subset of OR-Bench~\citep{cui2024orbench}. During curation, we assign 177 easy, 165 medium, and 158 hard items based on the reasoning each item requires, without using model predictions or independent calibration. For example, a refusal decision can involve a direct request or one whose intent is disguised by context, while a policy judgment can require a direct rule match or depend on multi-hop references and interacting exceptions. The decision format varies by source. In RAGTruth, the model either judges whether a response contains a hallucination relative to the supplied material (90 items) or identifies which segment, if any, contains the sole hallucinated span (60). ReasoningShield items ask the model to classify a reasoning trace as harmful or safe (45) or assign one of three risk levels (30), using human majority labels as references. HarmThoughts items ask the model to classify a target step into one of 16 behavior categories. DynaBench items ask either whether a dialogue complies with a given policy (50) or which rule, if any, it violates (50). Aegis items assess prompt safety (50) or response safety (50), and OR-Bench items test over-refusal on hard benign prompts. We score each decision against its reference label. Of the 500 references, 100 are model-derived, comprising 50 original HarmThoughts LLM annotations, 25 Aegis response labels from an LLM jury, and 25 OR-Bench labels based on hard-benign subset membership.

\subsection{Evidence Assessment}
\label{app:construction-evidence}

\paragraph{Retrieval \& Reranking (300).}
We construct 300 pairs from four sources. From TREC Deep Learning 2019/2020~\citep{craswell2020trecdl,craswell2021trecdl}, we sample 100 query--passage pairs, split evenly between the endpoint grades of perfectly relevant and irrelevant. From BRIGHT~\citep{su2024bright}, we select 70 pairs across biology, economics, sustainable living, and earth science. Of these, 35 are relevant, and 35 pair a query with a non-gold corpus passage retrieved by BM25 and screened with LLM assistance. From Touch\'e 2020~\citep{bondarenko2020overview,thakur2021beir}, we sample 30 pairs across ten topics, with one pair per topic at each native grade from 0 to 2. From ACORD~\citep{wang2025acord}, we sample 100 requirement--clause pairs, with 25 per grade from 1 to 4 on the native 0--4 scale. We assign TREC to the easy tier, BRIGHT and Touch\'e to the medium tier, and ACORD to the hard tier. Each item presents a single pair and asks for a pointwise relevance judgment. We score predictions against the reference judgments and award partial credit for predictions one grade away from the reference on ACORD. Pairs that share a query are statistically dependent.

\paragraph{Task Completion Verification (200).}
We construct 200 items from five sources. For BrowseComp-Plus~\citep{chen2025browsecompplus}, we define new subgoals and use snippets from a fixed historical search run as evidence. Each of 25 search prefixes yields one fully supported and one partially supported item, for a total of 50. We also include 8 CALVIN~\citep{mees2022calvin} light-state checks on official debug states, 56 $\tau$-bench~\citep{yao2025taubench} checks of backend records after replaying historical tool calls, and 46 MiniGrid~\citep{chevalierboisvert2023minigrid} state-predicate checks on newly collected seeded rollouts. The Embodied Agent Interface~\citep{li2024embodied} supplies the base tasks for 40 VirtualHome items. We assign difficulty by the minimum information and operations needed for a decision. Easy items require a direct check of the current state, medium items require integrating a bounded set of records or evidence, and hard items require inspecting the execution history. The VirtualHome subset contains 10 easy and 10 medium current-state checks and 20 hard history-dependent items. Each hard item starts with a successful execution of one of five base tasks. We append an executable sequence, called the audit suffix, that changes appliance states, removes two target objects, and restores zero, one, or both. A trajectory can therefore end with the objects in their required positions even if they were misplaced when an appliance was operated. The model judges three conditions. C1 requires at least one switch-on in the suffix and checks that all required object placements hold immediately before every such action. C2 requires at least one close action in the suffix and checks that all required placements hold immediately before the first close. C3 checks whether all original state and placement goals hold after the last action. The model selects one of four three-bit profiles, and each distractor flips exactly one bit of the correct profile. For each base task, we generate and verify all eight profiles, retaining four for evaluation and four as counterfactual witnesses. For every evaluated trajectory, a verified counterfactual reaches the same final goal state while changing the value of C1 or C2. Each condition is true in 10 of the 20 evaluated items. We use the verified profiles as references for the hard items and new model-derived annotations as references for the BrowseComp-Plus items. For evidence-sufficiency questions, a negative answer indicates that the visible evidence is insufficient to establish the requirement. We use the base task as the resampling unit when estimating uncertainty for the 20 hard items.

\subsection{Agent Task Completion}
\label{app:construction-agent}

\paragraph{Tool \& API Use (282).}
We construct 282 episodes from 204 BFCL~\citep{patil2025bfcl} requests and 78 ToolSandbox~\citep{lu-etal-2025-toolsandbox} scenarios. The easy tier contains 100 BFCL requests that specify an operation explicitly or require a direct lookup. Of these, 26 start at the first turn and 74 start at a later turn. The medium tier contains all 78 ToolSandbox scenarios, each presented as a single self-contained request. The hard tier contains 104 original first-turn requests from multi-turn BFCL tasks. Across the tier, these requests involve dependent lookups, conditional actions, or several coordinated outcomes. For episodes that start partway through an interaction, we replay the earlier BFCL turns or execute the earlier ToolSandbox turns in the simulator and show the resulting history to the model. We automatically resolve decisions with a single available option. Each episode has separate budgets for model decisions and native tool calls, with limits of 96 decisions and 12 calls for hard BFCL tasks. Tool calls that exceed time, memory, or output limits return an error and have no effect. \Cref{app:tool-scoring} describes how the episodes are scored.

\paragraph{Text-based Web Interaction (337).}
We construct 337 items from 100 single steps of Mind2Web~\citep{deng2023mind2web} demonstrations and 237 BreakingWeb~\citep{yin2026breakingweb} tasks across seven simulated applications. The Mind2Web steps form the easy tier and test whether the model can identify the relevant page element given the operation and any text to enter. The medium and hard tiers contain 137 and 100 BreakingWeb tasks, with 361 and 969 scored decisions, respectively. To enter the hard tier, a task must have at least five scored decisions or at least three entity choices whose targets are not named in the instruction. It must also include at least one decision that requires recalling a value from an earlier page, choosing from a menu with more than 60 options, or inferring several entities. In the easy tier, 50 items ask whether a supplied action advances the goal and 50 ask the model to select the best of four actions, using other controls on the same page as distractors. In the medium and hard tiers, the model makes decisions about which entity to act on, what exact value to enter, and when to stop. To fill a field, the model first selects the field, then selects a group if there are more than 64 candidate values, and finally selects the exact text span to enter. Reference routes are found by an external LLM or written by hand, and we replay each route to successful completion under the original checker. We also accept alternative action orders and detours when replay confirms that the task can still be completed.

\subsection{Scoring of Tool-Use Episodes}
\label{app:tool-scoring}

Each tool-use episode is scored by the checker of its source benchmark.

\paragraph{BFCL.} An episode scores 1 if BFCL's multi-turn checker accepts it and 0 otherwise. The checker compares the state of the simulated backend and the results of the model's tool calls with those of the reference calls~\citep{patil2025bfcl}.

\paragraph{ToolSandbox.} ToolSandbox specifies each scenario by milestones, events that must occur in order, and minefields, events that must not occur. Its native score measures how closely the episode matches the milestones and is 0 if any minefield occurs~\citep{lu-etal-2025-toolsandbox}. We make two adaptations. We omit the wording checks on final acknowledgements, because our interface accepts only structured decisions, and we award a score of 1 whenever the full goal state is reached without a minefield violation.

\paragraph{Exhausted budgets.} Episodes that exhaust their decision or tool-call budget (\Cref{app:construction-agent}) are scored in the same way.

\begin{table}[!htbp]
  \centering
  \captionsetup{font=footnotesize}
  \caption{\textbf{Releases and settings.} Open-weight models: the Hugging Face repository at the revision used, and the inference path of our latency and reliability runs, which follows each developer's documented fast path. APIs: the model identifier and settings of the evaluation runs.}
  \label{tab:config-settings}
  \fontsize{7.6}{9.1}\selectfont
  \setlength{\tabcolsep}{4pt}
  \begin{tabularx}{\linewidth}{@{}l>{\raggedright\arraybackslash}p{2.45in}>{\raggedright\arraybackslash}X@{}}
    \toprule
    Model & Release & Settings \\
    \midrule
    GLiDE & \nolinkurl{fastino/GLiDE} @\,d199b6d & In process (\texttt{Glide.decide}), BF16, SDPA attention, flash-linear-attention \\
    KnowLine-31B & \nolinkurl{PelaAI/KnowLine-31B-Gen1} @\,8f052a8 & SGLang 0.5.21, BF16 with an FP32 output layer, the documented read-twice front end \\
    Sieve-31B-v3 & \nolinkurl{sthanika-ai/Sieve-31B-v3} @\,44aa517 & In process (\texttt{load\_sieve}), BF16, eager; the released policy, which scores each request in two option orders in one pass \\
    Decider-27B & \nolinkurl{perplexity-ai/pplx-decider-v1.2} @\,cce0f3f & Documented fast path (layer compilation, CUDA graphs, length buckets) with FP8 matrix multiplication \\
    Kev-27B & \nolinkurl{jaredpalmer/kev-27b} @\,af0e6d5 & Documented serving defaults in process: BF16, fused kernels, CUDA graphs \\
    JevAny-27B & \nolinkurl{SimpleJev/JevAny-Qwen3.8-27B-LoRA} @\,09c9e91 on \nolinkurl{Qwen/Qwen3.8-27B} @\,1d4bf0f & Adapter merged in BF16, SDPA attention, flash-linear-attention, CUDA graphs \\
    StartLux Auto & \nolinkurl{startlux-models/StartLux-Decision-v1.1-4B-AutoThink} @\,23dd4c5 & Documented server on vLLM 0.19.1 with FP8; auto thinking with gate 0.6 and at most 4{,}096 thinking tokens \\
    StartLux Instant & \nolinkurl{startlux-models/StartLux-Decision-v1.1-4B-Instant} @\,416088f & Documented server on vLLM 0.19.1 with FP8 \\
    Clef-27B & \nolinkurl{Cloudflare/clef} @\,a30162b & Documented SGLang build (\texttt{dev-clef}) \\
    H2O Lightning & \nolinkurl{h2oai/h2o-lightning-4b} @\,08e1da5 & The developer's vLLM 0.30.0 build with FP8 \\
    Clef-Flash-9B & \nolinkurl{Cloudflare/clef-flash} @\,f0b7d3c & As Clef-27B \\
    Decider Flash & \nolinkurl{perplexity-ai/pplx-decider-flash-v1.2} @\,9ff7cf9 & Documented fast path, BF16 \\
    JevAny-4B & \nolinkurl{SimpleJev/JevAny-Qwen3.5-4B-LoRA} @\,1c7aa9b on \nolinkurl{Qwen/Qwen3.5-4B} @\,851bf6e & As JevAny-27B \\
    SII-4B & \nolinkurl{jylin001206/SII-Decision-4B-LoRA} @\,5c77219 & \texttt{DecisionEngine}, BF16; flash-linear-attention added in our runs (SII-4B + fla) \\
    Kev-4B & \nolinkurl{jaredpalmer/kev-4b} @\,6cfce5c on \nolinkurl{Qwen/Qwen3.5-4B-Base} @\,1001bb4 & As Kev-27B, with the adapter merged \\
    Jeva-4B & \nolinkurl{LIQIIIII/Jeva-4B} @\,f78e14e & \texttt{predict()} with the documented defaults; option cache warm or cold (\Cref{app:latency}) \\
    Laya & \nolinkurl{convaiinnovations/laya} @\,7b928d8 & TileLang fast path, 8{,}192-token input window; options not cut at 48 tokens \\
    \midrule
    GPT-6 Sol & \texttt{gpt-6-sol} & Responses API, medium reasoning effort, at most 25{,}000 output tokens \\
    GPT-6 Luna & \texttt{gpt-6-luna} & As GPT-6 Sol \\
    Jev API 1.14 & \texttt{jev-1.14.0} & TypeSafe decision API \\
    OpenAI Decisions & \texttt{gpt-6-luna} & OpenAI decision API \\
    \bottomrule
  \end{tabularx}
\end{table}
\FloatBarrier

\Needspace{5\baselineskip}
\section{Model Details}
\label{app:configs}

\Cref{tab:config-settings} lists the release of each open-weight model with the settings of our latency and reliability runs, and the model identifier and settings of each API.

\paragraph{Exceptions to the documented settings.}
Laya's official package cuts each option text at 48 tokens; all runs use the package without this cut, so that every option is read in full. In our latency and reliability runs, SII-4B + fla adds flash-linear-attention to the documented setup of SII-4B; its leaderboard score is that of the documented setup.

\paragraph{Reliability runs.}
Each open-weight System One model ran in the settings of the latency measurements (\Cref{app:latency}), one request at a time per model copy, on NVIDIA H100 80\,GB GPUs; the two System One APIs ran with four requests in flight. Eight of the 18 corrected references belong to the reliability sets. In the option-order test (six of the eight), we rescored the existing predictions; in the null-awareness test (all eight), we reran the affected items with the corrected option removed. Jeva-4B and StartLux Instant also expose their own null output, and we ran the null-awareness test a second time with it switched on, in the original choice format. For each of the two, $N$ uses the format that separates the two versions better on the 597 items: the own null output for StartLux Instant and the yes/no question for Jeva-4B. The comparison of StartLux Instant with larger models in \Cref{sec:exp-reliability} therefore rests on its own null output. Jeva-4B encodes each option separately, so the order of the options cannot change its scores: its $S$ is 1 by construction.

\paragraph{Unanswered requests and scoring.}
\label{app:zero-credit}
We distinguish three cases. \textit{(1) Context-window limits.} In 30 Web tasks, at least one request exceeds Laya's 8{,}192-token window. We retain these tasks without truncating their inputs and assign zero task credit under the context-window rule. \textit{(2) Static-task and tool-episode refusals.} OpenAI Decisions refuses four static tasks and twelve environment-based episodes. The Safety requests contain harmful traces and instructions to bypass safety restrictions, which may trigger refusal even though the evaluation asks only for classification (\Cref{fig:refusal-case-traces}). An explicit refusal leaves a required decision unanswered, so the affected static task or tool episode receives zero credit under the refusal rule. \textit{(3) Web-step refusals.} OpenAI Decisions refuses six exact-value requests in Web tasks involving a learning-management system, a patient portal, and Reddit. Potential triggers include sensitive page content; for example, the patient-portal request includes diagnoses and immunization records while asking the model to select a visit reason (\Cref{fig:refusal-web-structure}). Each refused decision is scored as incorrect, and route progress stops at the first incorrect decision. Credit for the preceding correct decisions is retained.

\paragraph{Scored runs and our runs.}
The scores of \Cref{sec:exp-main} come from runs separate from those behind the latency and reliability results. On the 939 option-order items in their original order, the two give the same credit on 907 to 938 items for each open-weight model except StartLux Auto, which samples its thoughts and agrees on 875. For SII-4B, the comparison is between its documented setup (scored) and SII-4B + fla (ours).

\clearpage
\begin{figure}[p]
  \centering
  \begin{adjustbox}{max totalheight=0.88\textheight}
    \begin{minipage}{\linewidth}
      \centering
      \input{figures-src/refusal-safety-structure}
    \end{minipage}
  \end{adjustbox}
  \captionsetup{font=footnotesize}
  \caption{\textbf{A Safety classification refusal.} The supplied context combines instructions to disregard safety restrictions with a harmful request, while the evaluation asks only for classification of a reasoning step. OpenAI Decisions returns a refusal rather than a category, yielding zero task credit.}
  \label{fig:refusal-case-traces}
\end{figure}

\begin{figure}[!p]
  \centering
  \begin{adjustbox}{max totalheight=0.84\textheight}
    \begin{minipage}{\linewidth}
      \centering
      \input{figures-src/refusal-web-structure}
    \end{minipage}
  \end{adjustbox}
  \captionsetup{font=footnotesize}
  \caption{\textbf{A refusal on a Web appointment task.} The request asks for an exact value from the task or page to fill a visit-reason field; the available context includes diagnoses, immunizations, and overdue-status information. OpenAI Decisions refuses this selection. The first three scored decisions retain $3/17=17.65\%$ progress; the fourth includes both a wrong group choice and the refusal.}
  \label{fig:refusal-web-structure}
\end{figure}
\clearpage

\Needspace{5\baselineskip}
\section{Latency Measurement}
\label{app:latency}

\paragraph{Tasks.}
The 1{,}109 timed tasks are every single-question choice task of seven static domains, across all tiers: Math 200, Scientific Reasoning 200, Knowledge \& Commonsense 200, Safety 190, Sentiment \& Emotion Understanding 150, Task Completion Verification 122, and Intent \& Routing 47.

\paragraph{Model settings.}
Each model runs the fast path documented by its authors (\Cref{tab:config-settings}), at their settings for precision, fused kernels, and CUDA graphs, and uses FP8 only where its documentation specifies it. Jeva-4B caches option encodings across requests; we time it with the cache kept within a pass (warm) and with the cache cleared before every request (cold). \Cref{fig:latency} shows the warm time.

\paragraph{Timing.}
Each model runs on one NVIDIA H100 80\,GB GPU used by no other process, with one request per call. The clock starts when the request is handed to the model's documented entry point and stops when the option probabilities are back on the host; it therefore includes tokenization and, for server-based models, the local HTTP round trip. After one untimed warm-up pass over all tasks, each model makes three timed passes, and a task's latency is the mean of its three timings. Two copies of each model ran at the same time on two GPUs, each on its own half of the tasks and pinned to its own 24 CPU cores.

\paragraph{Results.}
\Cref{tab:latency} lists the latency of every timed model.

\begin{table}[h]
  \centering
  \captionsetup{font=footnotesize}
  \caption{\textbf{Latency per task (ms)} on the 1{,}109 timed tasks: mean, 90th and 99th percentiles over tasks. Jeva-4B is timed with a warm and a cold option cache.}
  \label{tab:latency}
  \fontsize{7.6}{9.1}\selectfont
  \begin{tabular}{@{}lrrrr@{}}
\toprule
Model & Size & Mean & 90th percentile & 99th percentile \\
\midrule
Laya & 0.4B & 6.0 & 10.2 & 37.2 \\
Kev-4B & 4B & 25.2 & 61.0 & 157 \\
JevAny-4B & 4B & 25.2 & 56.2 & 173 \\
H2O Lightning & 4B & 26.5 & 48.4 & 155 \\
StartLux Instant & 4B & 35.1 & 63.2 & 138 \\
Decider Flash & 4B & 39.8 & 39.2 & 149 \\
SII-4B + fla & 4B & 40.6 & 48.5 & 157 \\
Clef-Flash-9B & 9B & 53.5 & 69.7 & 230 \\
Jeva-4B & 4B & 60.9 & 72.9 & 142 \\
Jeva-4B, cold cache & 4B & 79.0 & 88.2 & 180 \\
Kev-27B & 27B & 90.3 & 181 & 598 \\
JevAny-27B & 27B & 93.9 & 233 & 697 \\
Clef-27B & 27B & 108 & 184 & 533 \\
GLiDE & 27B & 113 & 214 & 774 \\
Decider-27B & 27B & 153 & 145 & 584 \\
KnowLine-31B & 31B & 177 & 515 & 959 \\
Sieve-31B-v3 & 31B & 265 & 668 & 2{,}475 \\
StartLux Auto & 4B & 3{,}084 & 10{,}032 & 21{,}870 \\
\bottomrule
\end{tabular}

\end{table}

\paragraph{Closed APIs.}
The closed APIs cannot be timed on this hardware. \Cref{fig:latency} places them at their end-to-end times from the evaluation runs, which include the network and the provider's queue, with four requests in flight.

\Needspace{5\baselineskip}
\section{Analysis Details}
\label{app:analysis}

\paragraph{Task selection.}
The analysis of \Cref{sec:analysis-domains} uses the medium and hard tiers of each tiered domain (for Math, its 50 medium and 70 hard tasks) and all tasks of the two untiered domains: 2{,}061 tasks, with 109 to 323 per domain (\Cref{tab:analysis-domains}). The leaderboard selection (\Cref{eq:capability-score}) has as few as 20 tasks in one domain (Task Completion Verification).

\paragraph{Per-domain values.}
\Cref{tab:analysis-domains} lists the values plotted in \Cref{fig:analysis-domains}.

\paragraph{Thinking control.}
StartLux-4B-AutoThink ran in each of its three thinking modes on all 2{,}061 tasks: the 1{,}642 static tasks (8{,}167 questions per mode), sent as batched requests, and the 419 Tool \& API Use and Web Interaction tasks through the benchmark's runtime (5{,}500 to 5{,}800 model calls per mode). Our \emph{auto} run gives the same credit as the benchmark's own \emph{auto} run on 1{,}507 of the 1{,}642 static tasks (92\%) and on 376 of the 419 tool and web tasks (90\%); thinking is sampled, and our requests were batched.

\paragraph{Hard tiers only.}
\Cref{tab:analysis-hard-only} repeats the backbone and thinking gains on the hard tier of each tiered domain, together with all tasks of the two untiered domains (1{,}202 tasks, with Math's 70 hard tasks). The patterns of \Cref{sec:analysis-domains} are unchanged except in three domains. In Intent \& Routing, the backbone gain falls from 12.0 to 4.9 points. In Task Completion Verification, whose hard tier has 20 tasks, the backbone gain falls from 17.4 to 3.8 points and the thinking gain from 7.3 to $-15.0$. In Code Understanding, the backbone gain is 7.7 points on medium and hard combined and 7.1 on hard alone, and the thinking gain is 5.7 and 8.6.

\paragraph{Reliability against performance.}
In \Cref{fig:analysis-reliability}a and c, the trend is a quadratic fit of $S$ (or $N$) on the leaderboard score over the 16 System One models other than Laya and Jeva-4B. The shaded region extends the trend by a constant half-width, the 90th percentile of the absolute residuals, over scores from 40 to 74; its half-width is 0.07 for $S$ and 0.18 for $N$. For $N$, we estimate 95\% intervals by resampling the 597 items 2{,}000 times, with the same resample for every model. Each of Kev-27B, Jev API 1.14, and JevAny-27B exceeds each of the four highest (GLiDE, KnowLine-31B, Sieve-31B-v3, and Decider-27B), with the interval of the difference above zero in all 12 pairs.

\begin{table}[h]
  \centering
  \captionsetup{font=footnotesize}
  \caption{\textbf{Per-domain values behind \Cref{fig:analysis-domains}.} Tasks: medium and hard tiers, or all tasks of an untiered domain. Luna: GPT-6 Luna's score. Differences from Luna with 95\% bootstrap intervals over tasks. Backbone gain: mean over the four families. StartLux-4B-AutoThink: its scores in the three thinking modes.}
  \label{tab:analysis-domains}
  \fontsize{7.6}{9.1}\selectfont
  \setlength{\tabcolsep}{3pt}
  \begin{tabular}{@{}lrr cc r rrr@{}}
\toprule
 & & & \multicolumn{2}{c}{Difference from GPT-6 Luna} & Backbone & \multicolumn{3}{c}{StartLux-4B-AutoThink} \\
\cmidrule(lr){4-5}\cmidrule(lr){6-6}\cmidrule(l){7-9}
Domain & Tasks & Luna & GLiDE & StartLux Instant & gain & Never & Always & Auto \\
\midrule
Knowledge \& Commons. & 200 & 90.0 & +5.0 [+1.5, +9.0] & $-$1.5 [$-$6.0, +3.0] & +4.4 & 85.0 & 88.0 & 87.5 \\
Sentiment \& Emotion & 200 & 84.8 & +4.8 [+1.2, +8.5] & +3.0 [$-$2.0, +8.0] & +3.4 & 88.0 & 85.8 & 86.0 \\
Retrieval \& Reranking & 200 & 76.5 & $-$5.8 [$-$11.2, $-$0.2] & $-$8.2 [$-$14.5, $-$1.8] & +3.9 & 69.0 & 70.8 & 71.2 \\
\midrule
Intent \& Routing & 200 & 79.5 & +0.1 [$-$2.4, +2.9] & $-$8.9 [$-$12.0, $-$5.8] & +12.0 & 68.8 & 70.8 & 71.1 \\
Safety & 323 & 75.2 & +4.0 [$-$1.2, +9.0] & $-$7.1 [$-$12.7, $-$1.5] & +16.2 & 60.4 & 63.8 & 63.5 \\
Task Completion Verif. & 109 & 98.2 & +0.9 [+0.0, +2.8] & $-$22.9 [$-$31.2, $-$15.6] & +17.4 & 58.7 & 66.1 & 73.4 \\
Tool \& API Use & 182 & 75.8 & $-$3.6 [$-$9.1, +1.8] & $-$11.2 [$-$17.1, $-$5.2] & +25.6 & 61.9 & 64.5 & 61.9 \\
Web Interaction & 237 & 64.4 & +3.1 [$-$0.4, +6.5] & $-$28.8 [$-$33.8, $-$23.8] & +23.9 & 40.8 & 42.6 & 40.4 \\
\midrule
Math & 120 & 100.0 & $-$26.7 [$-$34.2, $-$19.2] & $-$38.3 [$-$47.5, $-$30.0] & +14.4 & 50.8 & 92.5 & 87.5 \\
Scientific Reasoning & 150 & 91.3 & $-$21.3 [$-$28.7, $-$14.7] & $-$27.3 [$-$35.3, $-$19.3] & +8.3 & 58.7 & 70.7 & 70.7 \\
Code Understanding & 140 & 88.6 & $-$25.7 [$-$34.3, $-$17.1] & $-$26.4 [$-$35.7, $-$17.1] & +7.7 & 55.7 & 61.4 & 59.3 \\
\bottomrule
\end{tabular}

\end{table}

\begin{table}[h]
  \centering
  \captionsetup{font=footnotesize}
  \caption{\textbf{Backbone and thinking gains on the medium and hard tiers and on the hard tier alone.} Gains as in \Cref{fig:analysis-domains}b and c (thinking: \emph{always} minus \emph{never}). The untiered domains are identical in both selections.}
  \label{tab:analysis-hard-only}
  \fontsize{7.6}{9.1}\selectfont
  \begin{tabular}{@{}l rrr rrr@{}}
\toprule
 & \multicolumn{3}{c}{Medium and hard tiers} & \multicolumn{3}{c}{Hard tier only} \\
\cmidrule(lr){2-4}\cmidrule(l){5-7}
Domain & Tasks & Backbone & Thinking & Tasks & Backbone & Thinking \\
\midrule
Knowledge \& Commons. & 200 & +4.4 & +3.0 & \multicolumn{3}{c}{untiered} \\
Sentiment \& Emotion & 200 & +3.4 & $-$2.2 & \multicolumn{3}{c}{untiered} \\
Retrieval \& Reranking & 200 & +3.9 & +1.8 & 100 & +3.4 & +4.0 \\
\midrule
Intent \& Routing & 200 & +12.0 & +2.0 & 100 & +4.9 & +3.6 \\
Safety & 323 & +16.2 & +3.4 & 158 & +19.8 & +4.4 \\
Task Completion Verif. & 109 & +17.4 & +7.3 & 20 & +3.8 & $-$15.0 \\
Tool \& API Use & 182 & +25.6 & +2.6 & 104 & +26.2 & +0.0 \\
Web Interaction & 237 & +23.9 & +1.8 & 100 & +21.5 & +2.7 \\
\midrule
Math & 120 & +14.4 & +41.7 & 70 & +13.6 & +52.9 \\
Scientific Reasoning & 150 & +8.3 & +12.0 & 80 & +11.2 & +13.8 \\
Code Understanding & 140 & +7.7 & +5.7 & 70 & +7.1 & +8.6 \\
\bottomrule
\end{tabular}

\end{table}

\end{document}